\documentclass[10pt,journal,compsoc]{IEEEtran}
\usepackage{amsmath}
\usepackage{amsfonts}
\usepackage{bbding}
\usepackage{amsmath}
\usepackage{amsfonts}
\usepackage[utf8]{inputenc} 
\usepackage[T1]{fontenc}    
\usepackage{hyperref}       
\usepackage{url}            
\usepackage{booktabs}       
\usepackage{amsfonts}       
\usepackage{nicefrac}       
\usepackage{microtype}      
\usepackage{amsthm,amsmath,amssymb}
\usepackage{mathrsfs}
\usepackage{graphicx}
\usepackage{subfigure}
\usepackage{color}
\usepackage{wrapfig}
\usepackage{bbding}
\usepackage{float}
\usepackage{multirow}

\ifCLASSOPTIONcompsoc
  \usepackage[nocompress]{cite}
\else
  \usepackage{cite}
\fi
\ifCLASSINFOpdf
\else
\fi
\begin{document}
%
\title{Learning in Audio-visual Context: \\
A Review, Analysis, and New Perspective}

%

%
%
%
%

\author{
        Yake Wei,
        Di Hu,
        Yapeng Tian,
        Xuelong Li,~\IEEEmembership{Fellow,~IEEE}
\IEEEcompsocitemizethanks{

\IEEEcompsocthanksitem Y. Wei and D. Hu (corresponding author) are with the Gaoling School of Artificial Intelligence, and Beijing Key Laboratory of Big Data Management and Analysis Methods, Renmin University of China, Beijing 100872, China.\protect\\
E-mail: \{yakewei, dihu\}@ruc.edu.cn

\IEEEcompsocthanksitem Y. Tian is with the Department of Computer Science, University of Texas at Dallas, TX 75080 Richardson, United States.\protect\\
E-mail: yapeng.tian@utdallas.edu

\IEEEcompsocthanksitem X. Li is with the School of Artificial Intelligence, OPtics and ElectroNics (iOPEN), Northwestern Polytechnical University, Xi’an 710072, P.R.China.\protect\\
E-mail: li@nwpu.edu.cn}
}

\IEEEtitleabstractindextext{%


\begin{abstract}
Sight and hearing are two senses that play a vital role in human communication and scene understanding. 
To mimic human perception ability, audio-visual learning, aimed at developing computational approaches to learn from both audio and visual modalities, has been a flourishing field in recent years.
A comprehensive survey that can systematically organize and analyze studies of the audio-visual field is expected.
Starting from the analysis of audio-visual cognition foundations, we introduce several key findings that have inspired our computational studies. Then, we systematically review the recent audio-visual learning studies and divide them into three categories:
audio-visual boosting, cross-modal perception and audio-visual collaboration. 
Through our analysis, we discover that, the consistency of audio-visual data across semantic, spatial and temporal support the above studies.
To revisit the current development of the audio-visual learning field from a more macro view, we further propose a new perspective on audio-visual scene understanding, then discuss and analyze the feasible future direction of the audio-visual learning area. 
Overall, this survey reviews and outlooks the current audio-visual learning field from different aspects. 
We hope it can provide researchers with a better understanding of this area. A website including constantly-updated survey is released: \url{https://gewu-lab.github.io/audio-visual-learning/}.
\end{abstract}

}

\maketitle

\IEEEdisplaynontitleabstractindextext

%
\IEEEpeerreviewmaketitle

\begin{figure*}[t]
    \centering
    \includegraphics[width=1\linewidth]{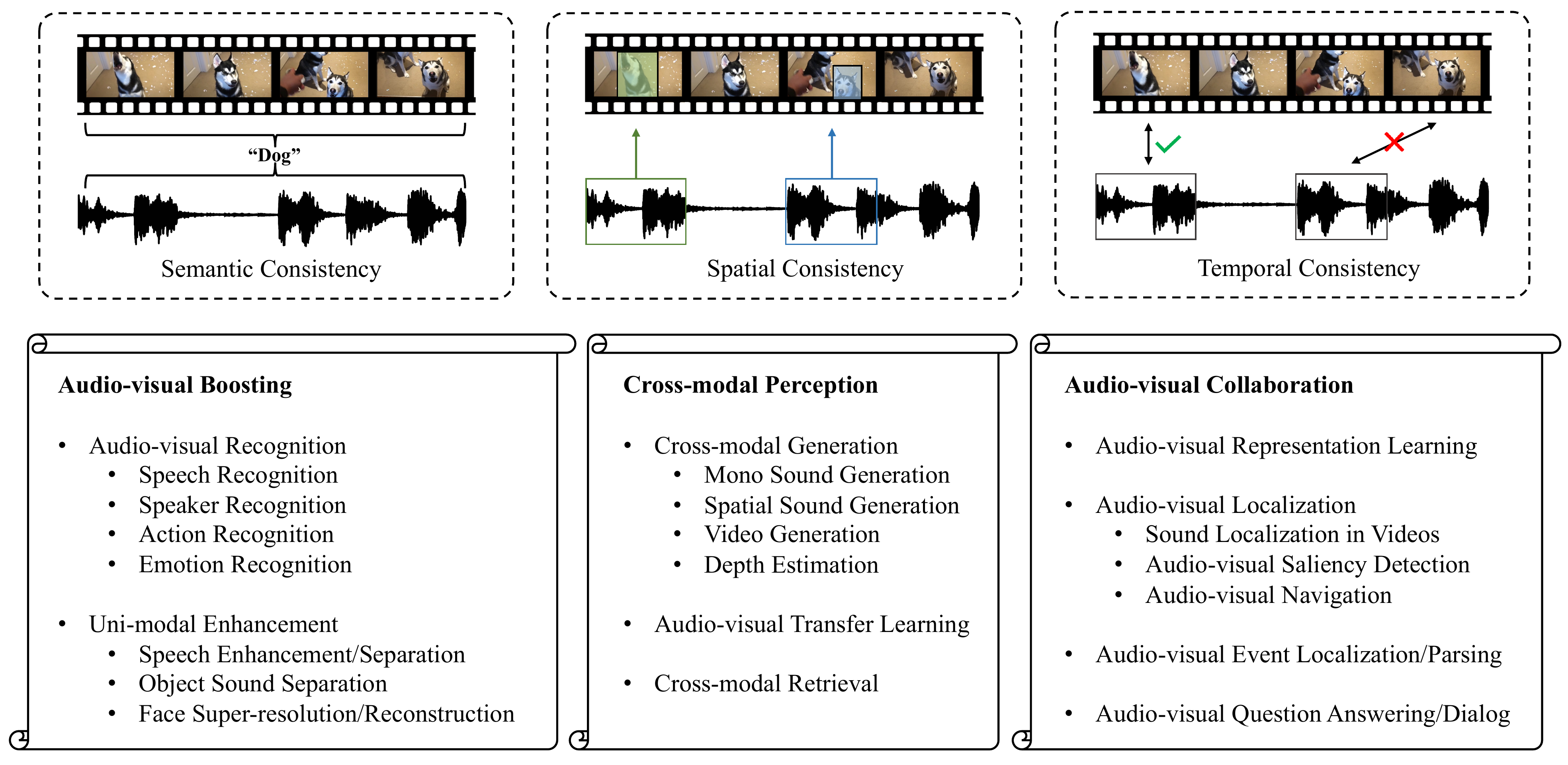}
    \vspace{-1.5em}
    \caption{\textbf{Illustration of multiple audio-visual consistencies and the overview of audio-visual learning field.} \textbf{Top:} The barking and appearance of a dog are both related to the semantic concept of ``dog'', and its spatial location can also be captured in visual or indicated by corresponding audio. Meanwhile, the audio-visual segments of the same timestamp are naturally consistent. \textbf{Bottom:} In this survey, we provide cognition-inspired categories and sub-categories to help organize existing audio-visual studies.}
    \vspace{-1em}
    \label{fig:consistency}
\end{figure*}

\section{Introduction}
\IEEEPARstart{S}{ight} and hearing, which mainly are relied on by humans when perceiving the world, occupy a huge portion of the received external information~\cite{man2018brain}. Our brain holds a comprehensive understanding of the environments via integrating these multi-modal signals with different forms and physical characteristics. For example, in the cocktail-party scenario with many speakers, we can locate the one of interest and enhance the received speech with the aid of his/her lip motion. Hence, audio-visual learning is essential to our pursuit of human-like machine perception ability. Its purpose is to explore computational approaches that learn from both audio-visual data. The special properties of audio-visual modalities make them distinct:

\textbf{1) Cognitive Foundation.} As two of the most widely studied senses, the integration of sight and hearing occurs broadly in the human nervous system, from the superior colliculus to the cortex~\cite{gazzaniga2010cognitive}. On the one hand, the importance of the two senses in human perception provides the cognitive basis for the study of machine perception based on audio-visual data, and on the other hand, the evidence that sight and hearing exhibit extensive interaction and integration, can facilitate the advance of audio-visual learning.

\textbf{2) Multiple Consistencies.} In our daily experience, sight and hearing are closely aligned. As illustrated in~\autoref{fig:consistency}, both the barking sound of the dog and its appearance allow us to relate to the concept of ``dog'' (\textit{Semantic Consistency}). Meanwhile, we can determine the exact spatial location of the dog with the help of either the heard sound or vision (\textit{Spatial Consistency}). And when hearing dog barking, we can usually see the dog at the same time (\textit{Temporal Consistency}). These diverse consistencies between sight and hearing build the basis of audio-visual learning studies.

\textbf{3) Abundant Data.}
The rapid development of mobile terminals and the Internet has inspired more and more people to share videos on public platforms, which makes it easier to collect videos at a lower cost. Since the high-quality large-scale dataset is the key infrastructure for developing effective audio-visual models, these abundant videos can alleviate the data acquisition obstacle.

The above properties make it natural to develop computational approaches to learn from both audio and visual modalities. We actually have witnessed the bloom of audio-visual learning in recent years. Especially, not just integrating an additional modality to improve existing uni-modal tasks, our community has begun to explore and solve new problems and challenges.
However, the existing audio-visual studies are often task-oriented. In these works, they typically focus on a specific audio-visual problem. The community still lacks a comprehensive survey that can systematically review and analyze works in this research field. To this end, we elaborate on the current audio-visual field, then further look over its shortcoming and outlook the future directions in this article. 
Since there is a tight connection between audio-visual learning and human perception ability, we first conclude the cognition foundation of audio-visual modalities, based on which, the existing audio-visual studies are divided into three categories as illustrated in~\autoref{fig:consistency}:

\textbf{1) Audio-visual Boosting.} 
The audio and visual data have been separately studied for a long time and there is a wide range of applications. Although methods of these uni-modal tasks mostly have achieved impressive results, they only take partial information about the observed things of interest, for which the performance is limited and vulnerable to uni-modal noise. Thus, researchers introduce data of another modality into these audio or visual tasks, which not only makes the model more robust but also boosts the performance via integrating complementary information.

\textbf{2) Cross-modal Perception.}
It is natural for humans to imagine the visual appearance or sound pattern based on the prior of another modality, employing the consistency between modalities. Correspondingly, the consistency between audio and visual data provides the possibility for machines to transfer learnt knowledge or generate modal-specific data according to information from another one. Therefore, many tasks are proposed to pursue this cross-modal perception ability, achieving significant results.

\textbf{3) Audio-visual Collaboration.}
Besides fusing signals of different senses, there is more high-level cooperation between multiple modalities in the cortical region for humans to achieve better scene understanding. 
Thus, the pursuit of human-like perception ability needs to explore the collaboration between audio-visual modalities. 
To reach this purpose, tasks that introduce new audio-visual scene understanding challenges, are proposed and widely focused on these years.

For all the above tasks, the audio-visual consistency across semantic, spatial and temporal provides the achievable foundation. Hence, we analyze these multiple consistencies after concluding the recent studies. Furthermore, to revisit the progress of the audio-visual learning field in human-like perception modelling, a new perspective on audio-visual scene understanding is proposed. Compared with existing surveys related to the audio-visual learning field in recent years~\cite{zhu2021deep,vilacca2022recent}, we start with the cognition foundation of audio-visual modalities, then the observed findings help us to systematically organize studies of the audio-visual field in a more logical paradigm, based on which the intrinsic consistency across modalities are also adequately discussed.
Specifically, Zhu \emph{et al.}~\cite{zhu2021deep} provided a survey of deep learning-based audio-visual applications. They only focused on some typical audio-visual tasks and lacked targeted analysis of audio-visual correlation, resulting in its limitation in scope and lacking in-depth discussion. Vila{\c{c}}a \emph{et al.}~\cite{vilacca2022recent} reviewed the recent deep learning models and their objective functions in audio-visual learning. They mainly paid attention to the techniques used in correlation learning, being short of a comprehensive review of the development in the whole field. In addition, several multi-modal or multi-view surveys~\cite{li2018survey,baltruvsaitis2018multimodal,chai2022deep} took audio-visual learning into consideration. But these works did not concentrate on this field in terms of the properties of audio-visual learning.

The rest of this survey is organised as follows. We start with a discussion about the cognitive foundation of audio-visual modality (\autoref{sec:foudation}), then conclude the recent progress of audio-visual tasks with three categories: Audio-visual Boosting (\autoref{sec:boost}), Cross-modal Perception (\autoref{sec:cross}) and Audio-visual Collaboration (\autoref{sec:both}). Afterwards,~\autoref{sec:dataset} surveys relevant audio-visual datasets for different tasks. Subsequently, the multiple consistencies of audio-visual modalities in the mentioned tasks and the new perspective about the scene understanding are further discussed in~\autoref{sec:discussion}. This survey is developed based on our hosted tutorial ``Audio-visual Scene Understanding'' in CVPR 2021. We hope it can offer researchers a comprehensive overview of the current audio-visual learning field.

\section{Audio-visual Cognitive Foundation}
\label{sec:foudation}

Sight and hearing are two core senses for human scene understanding~\cite{man2018brain}. In this section, we summarize the neural pathways and integration of audio-visual modality in cognitive neuroscience, to set the stage for the subsequent discussion on computational audio-visual studies.

\subsection{Neural Pathways of Audio and Visual Sense}
Vision is the most widely studied sense and thought to dominate human perception by some views~\cite{man2018brain}. Correspondingly, the neural pathway of vision is more complicated. Light reflected from objects contains visual information, and it activates the numerous photoreceptors (about $260$ million) on the retina. The output of photoreceptors is sent to only $2$ million ganglion cells. This process compresses the visual information, leaving it to be decoded by higher-level visual centers. Then, after being processed by lateral geniculate nucleus cells, visual information finally reaches higher-order visual areas in the cortex. The visual cortex is a combination of distinct regions with functional differences, whose visual neurons have preferences. For example, neurons in V4 and V5 are sensitive to color and motion, respectively~\cite{gazzaniga2010cognitive}.

Besides vision, audition is also an important sense to observe the surroundings. It not only can be essential for survival, which is helpful for the human to avoid possible attacks, but also is fundamental for communication~\cite{pannese2015subcortical}. Sound waves are transformed into neuronal signals at the eardrums. Then, the auditory information is transported to the inferior colliculus and cochlear nucleus in the brainstem. The sound is finally encoded at the primary auditory cortex, after the process of medial geniculate nucleus of the thalamus~\cite{gazzaniga2010cognitive}. The brain takes the auditory information, and then uses the acoustic cues embedded in them, such as frequency and timbre, to determine the identity of the sound source. Meanwhile, the intensity and interaural time differences between two ears provide cues for the sound location, which is called the binaural effect~\cite{merker2007consciousness}. In practice, human perception can combine diverse senses, especially audition and vision, which is called multi-sensory integration~\cite{gazzaniga2010cognitive}.

\subsection{Audio-visual Integration in Cognitive Neuroscience}
Each sense provides unique information about the surroundings. Although the received information of multiple senses is distinct, the resulting representation of the environment is a unified experience, rather than disjointed sensations. One representative demonstration is McGurk effect~\cite{mcgurk1976hearing}, where two different audio-visual speech content results in a single received information. In fact, human perception is a synthetic process, in which information from multiple senses is integrated. Especially, the intersection of neural pathways of audition and vision combines information from two vital human senses to facilitate perceptual sensitivity and accuracy, \emph{e.g.,} the sound-related visual information improves search efficiency in auditory space perception~\cite{jones1975eye}. These phenomena that the perception results combine the observation of more than one senses, have attracted attention in the cognitive neuroscience field. 

Based on existing studies, multi-sensory integration occurs brain regions ranging from subcortical to cortical~\cite{gazzaniga2010cognitive}. 
One well-studied subcortical multi-modal region is superior colliculus. Many neurons of the superior colliculus own multi-sensory properties, which can be activated by information from visual, auditory, and even tactile senses. This multi-sensory response tends to be stronger than a single one~\cite{stein2004crossmodal}. The superior temporal sulcus in cortex is another representative region. It has been observed to connect with multiple sensory based on the studies of monkeys, including visual, auditory, and somatosensory~\cite{hikosaka1988polysensory}. More brain regions, including the parietal, frontal lobes and hippocampus, show a similar multi-sensory integration phenomenon. 

According to these studies about multi-sensory integration, we can observe several key findings: \textbf{1) Multi-modal boosting.} As stated above, many neurons can respond to the fused signals of multiple senses and such an enhanced response is more reliable than the uni-modal one when the stimulus of a single sense is weak~\cite{stein2004crossmodal}. \textbf{2) Cross-modal plasticity.} This phenomenon refers that the deprivation of one modality is able to impact the cortex development of the remaining modalities. For example, the auditory-related cortex of the deaf has the potential to be activated by visual stimulate~\cite{cohen1997functional}. \textbf{3) Multi-modal collaboration.} Signals of different senses have more complex integration in cortical regions. Researchers found that the cortex forms a module with the capacity of integrating multi-sensory information in a collaborative manner to build awareness and cognition~\cite{zamora2010cortical}. Eventually, the audio-visual integration, occurs broadly in cognition, helping us own more sensitive and accurate perception with the support of the above finding.

Inspired by the cognition of humans, researchers have started to investigate how to achieve human-like audio-visual perception ability. For example, synaesthesia is the phenomenon that information from multiple senses is united in an idiosyncratic manner. Li \emph{et al.}~\cite{li2008visual} first viewed this phenomenon as theory basis and proposed to bridge the content association between image and music. Without loss of generality, numerous audio-visual studies have emerged in recent years.

\begin{figure*}[t]
    \centering
    \includegraphics[width=1\linewidth]{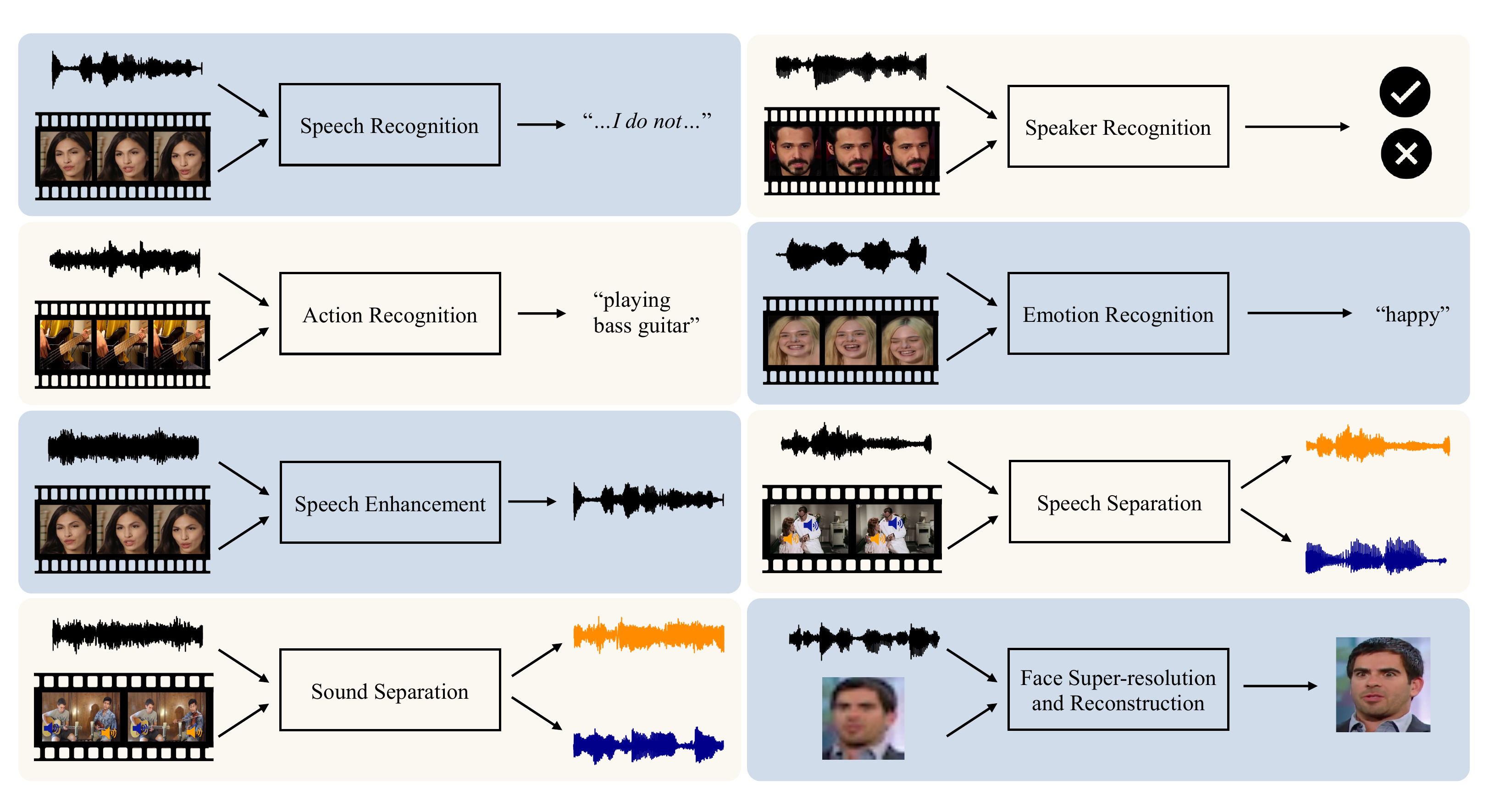}
    \vspace{-2em}
    \caption{\textbf{Illustration of the audio-visual boosting tasks.} The audio-visual boosting tasks include audio-visual recognition and uni-modal enhancement. These tasks aim to boost the performance of existing uni-modal tasks by introducing additional modality, based on the audio-visual consistency across semantic, spatial and temporal.}
    \vspace{-1em}
    \label{fig:boost}
\end{figure*}

\section{Audio-visual Boosting}
\label{sec:boost}
Audio-visual data depict things of interest from different views. Even though individual modalities themselves already contain a rich amount of information for the machine to utilize, they only observe partial information and are sensitive to uni-modal noise, which limits the performance of uni-modal methods. Inspired by the multi-modal boosting cases of human cognition, audio-visual modalities are both introduced to boost task performance. In~\autoref{fig:boost}, we provide the sketches of relevant tasks.

\subsection{Audio-Visual Recognition}
Uni-modal recognition tasks have been widely studied in the past, like audio-based speech recognition~\cite{reddy1976speech} and visual-based action recognition~\cite{poppe2010survey}. However, since the uni-modal methods only observe partial information about the things of interest, which is vulnerable to uni-modal noise, audio-visual recognition that fuses multi-modal information according to their consistencies, has raised the attention over the years to boost the robustness and capacity of models.

\textbf{Speech Recognition.} Audio speech recognition task has been studied for many years~\cite{huang2014historical}, which aims to translate audio information into format text. However, the audio quality is sensitive to the channel, environment and speech style in practice, making speech recognition performance degrade greatly. To further obtain the robust model, researchers are inspired by human speech perception that relies on both audio and visual to decide what is being said, especially in noisy scenarios. Concretely, the movement of the speech-related face region is often synchronized with the corresponding audio temporally. Meanwhile, the revealed speech content of audio and face, especially lips, is also highly consistent. Thus, stable visual information in various acoustical environments is introduced to build a more robust speech recognition system. In early studies, the visual features are extracted by prior lip-shape representation frameworks~\cite{luettin1996visual,chen2001audiovisual} or some low-level visual algorithms~\cite{matthews2001comparison}. Then, \emph{Hidden Markov Model} (HMM)-related methods are generally utilized in the modelling of audio-visual synchrony state over time~\cite{dupont2000audio,nefian2002dynamic,gurban2008dynamic}. Subsequently, the rising neural networks are utilized to extract features~\cite{noda2015audio}, since the hand-crafted features either require fine labelling or are vulnerable to factors such as illumination. What's more, the sequential information of videos can be better modelled via \emph{Recurrent Neural Networks} (RNNs)~\cite{petridis2017end,hu2016temporal}. In these years, the transformer-based recognition framework is introduced to further build context dependence as well as cross-modal association~\cite{afouras2018deep,zhao2019explicit,song2022multimodal}. These novel model architectures bring the ability of audio-visual speech recognition to a new peak, while the dependence of deep learning methods on large-scale available data, makes the cost of labelling a huge burden. Hence, Shi \emph{et al.}~\cite{shi2022robust} attempted to build the audio-visual speech recognition model with unlabeled data via the clustering algorithm. Generally, after introducing the visual modality, the speech recognition methods have evolved considerably in performance.

\textbf{Speaker Recognition.} Similar to speech recognition, speaker recognition also focuses on the speech scenarios. It is a series of tasks that pay attention to the identity of speaker, including speaker verification, identification and diarization. Speaker verification aims to verify whether an utterance is pronounced by a target speaker and speaker identification aims to determine speaker identity for an utterance. Speaker diarization answers the question ``who spoke when''. Audio and facial information are two important personal identities and speaker recognition methods with one of which have been a mature field~\cite{jafri2009survey,reynolds2002overview}. But these uni-modal methods are vulnerable to the variants of environments. The acoustic characteristics of audio are susceptible to many factors including background noise and the speaker's mood, while the facial information is sensitive to pose, illumination, and device. Therefore, the combination of multiple modalities is essential for building a more robust model, considering personal identities are semantically consistent in both audio and visual. In the early studies of speaker verification and identification, researchers often utilized hand-crafted features and conducted decision fusion to combine audio expert and visual expert~\cite{ben1999fusion,choudhury1999multimodal,wu2006multi,sargin2009audiovisual}. As deep learning develops by leaps and bounds, the uni-modal features tend to be extracted by neural networks and more feature-level fusion methods emerge, like concatenation~\cite{sari2021multi} and attention mechanism~\cite{shon2019noise,qian2021audio}, besides decision fusion~\cite{sell2018audio,sari2021multi}. Compared with speaker verification and identification, the speaker diarization task needs to further split the given speech into speaker-homogeneous segments. After introducing visual modality, studies of diarization include two paths. One path is to enhance diarization performance via fusing uni-modal results~\cite{sarafianos2016audio,friedland2009multi,noulas2011multimodal}. Another path extends the diarization task into ``who spoken when and from where'' that further localizes the speech-related region in visual~\cite{vajaria2006audio,friedland2009visual,gebru2017audio,chung2019said}. In recent years, the requirement for numerous data of deep learning has resulted in the burden of collecting as well as labelling diarization datasets. Thus, the nature of audio-visual synchronization is further utilized to design the self-supervised audio-visual diarization method~\cite{ding2020self}. Overall, audio-visual speaker recognition achieves impressive results with information from multiple views.

\textbf{Action Recognition.} In the computer vision field, action recognition is a widely studied task, and it aims to discriminate different human activities based on the video. Researchers have employed many visually relevant modalities for action recognition~\cite{sun2022human}. For example, depth modality owns spatial information and skeleton modality reveals structural information of subject pose. But these modalities are all limited to reflecting visual information although with different emphases. In contrast, the audio modality is promising to serve as complementary information in semantics. Firstly, plenty of action categories company with sound. Secondly, audio can reflect the discriminative information of similar actions in visual. For example, the actions of \textit{eatboxing} and \textit{singing} in Kinetics-700~\cite{carreira2019short}.
Thirdly, audio can capture the out-of-screen or visually hard to observe actions (\emph{e.g.,} \textit{sniffing}). 
Thus, audio-visual action recognition is introduced to perceive actions with more views. Among these studies, decision fusion and late fusion are commonly used to combine the audio-visual information~\cite{wang2016exploring,ghanem2018activitynet}. More recently, more fine-grained mid-level fusion methods are proposed to fully explore the audio-visual consistency in temporal~\cite{kazakos2019epic,xiao2020audiovisual}. Kazakos \emph{et al.}~\cite{kazakos2019epic} presented the Temporal Binding Network to aggregate RGB, Flow and Audio information temporally at mid-level. In these years, the flourish of transformer~\cite{vaswani2017attention} has inspired the transformer-based mechanism that both considers self-attention and cross-modal attention~\cite{chen2022mm}. What's more, since videos often contain action-irrelevant segments, resulting in unnecessary computation cost and even possibly interfering with the prediction, audio modality is used to reduce redundancy segments of videos~\cite{korbar2019scsampler,gao2020listen}. Similarly, the modality selection~\cite{Panda_2021_ICCV} and dropout~\cite{alfasly2022learnable} strategies are adopted for efficient video recognition. Apart from the above methods that fusing multiple information, some researchers consider the audio modality as the assistance for the domain generalization in visual, alleviating the domain shift problem in action recognition across scenes~\cite{planamente2021cross,planamente2022domain,zhang2022audio}. Overall, the introducing of audio enhances the action recognition task from different perspectives, including richer information, more efficient training, and better generalization.

\textbf{Emotion Recognition.} Compared to the above recognition tasks that have clear classification criteria, the emotion recognition task is more difficult since the boundary between sentiments is even ambiguous for humans. The emotion recognition task pays attention to recognizing the sentiment of humans, which can be reflected in various aspects, including facial expression, audio, gestures even body movement. Based on~\cite{mehrabian2017communication}, facial expression ($55\%$) and vocal ($38\%$) contribute most of the human sentiment. Therefore, audio-visual input is important for the emotion recognition task. At the early stage of the emotion recognition study, features of multiple modalities are generated by hand-crafted technique. The audio features are often extracted based on the acoustic characteristic, like pitch and energy~\cite{sebe2006emotion,lin2011error,wu2013two}, while visual features are often based on facial texture (\emph{e.g.,} wrinkles, furrows) or components (\emph{e.g.,} eyebrows, mouth, eyes)~\cite{pantic2007machine,fasel2003automatic}. The fusion operations of these studies can be categorized into feature-level fusion~\cite{metallinou2012context,eyben2012audiovisual,rudovic2013bimodal}, decision-level fusion~\cite{metallinou2010decision,ramirez2011modeling} and model-level fusion~\cite{zeng2008audio,jiang2011audio}. However, the hand-crafted features rely on expert knowledge to design effective extraction methods. Even though, they are still with limited representation capability, and have difficulties in exploring multiple inter- and intra-modal correlations. More recently, the more powerful deep neural network is widely employed in the emotion recognition task, and more diverse inter-, as well as intra-modal fusion strategies are proposed~\cite{zadeh2017tensor,zadeh2018memory,hazarika2018conversational,Lv_2021_CVPR}. Also, the transformer-based framework is introduced since its advantage in capturing global and local attention~\cite{delbrouck2020transformer}. Besides the above conventional emotion recognition, the sentiments with below-the-surface semantics, like sarcasm or humour, have also begun to consider multi-modal signals including visual expression and speech patterns for better capturing the crucial sentiment cues~\cite{chauhan2020sentiment,bedi2021multi}.

\begin{table*}[!t]
\renewcommand\arraystretch{1.3}
\centering
\caption{A summary of audio-visual boosting tasks.}
\vspace{-1em}
\label{tab:boost}
\setlength{\tabcolsep}{1mm}{
\begin{tabular}{cccc}
\toprule
Task                                                                                & Problem of uni-modal input                                                                                                                 & Motivation of audio-visual input                                                                                                   & Representative methods                                                                                                                                                                               \\ \toprule
Speech recognition                                                                  & Audio is sensitive to acoustic noise.                                                                                                      & \begin{tabular}[c]{@{}c@{}}Visual can indicate speech content,\\ and is steady in different acoustic scenes.\end{tabular}          & \begin{tabular}[c]{@{}c@{}}\cite{noda2015audio,hu2016temporal,petridis2017end}\\ \cite{afouras2018deep,zhao2019explicit,song2022multimodal,shi2022robust}\end{tabular}                                   \\ \cline{2-4} 
Speaker recognition                                                                 & \begin{tabular}[c]{@{}c@{}}Uni-modal information is vulnerable \\ to the variants of environments.\end{tabular}                            & \begin{tabular}[c]{@{}c@{}}Both audio and visual contain consistent \\ personal identities information.\end{tabular}               & \begin{tabular}[c]{@{}c@{}}\cite{sell2018audio,chung2019said,shon2019noise}\\ \cite{ding2020self,sari2021multi,qian2021audio}\end{tabular}                                                           \\ \cline{2-4} 
Action recognition                                                                  & \begin{tabular}[c]{@{}c@{}}Vision information is limited \\ in the description of actions.\end{tabular}                                    & \begin{tabular}[c]{@{}c@{}}Audio can provide complementary\\ action-related information.\end{tabular}                              & \begin{tabular}[c]{@{}c@{}}\cite{wang2016exploring,ghanem2018activitynet,kazakos2019epic}\\ \cite{korbar2019scsampler,xiao2020audiovisual,gao2020listen,Panda_2021_ICCV}\end{tabular}                \\ \cline{2-4} 
Emotion recognition                                                                 & \begin{tabular}[c]{@{}c@{}}Uni-modal information is not \\ comprehensive to represent emotion.\end{tabular}                                & \begin{tabular}[c]{@{}c@{}}Facial expression and vocal\\ contribute most to human sentiment.\end{tabular}                          & \begin{tabular}[c]{@{}c@{}}\cite{zadeh2017tensor,zadeh2018memory,hazarika2018conversational,ghosal2018contextual}\\ \cite{cai2019multi,shenoy2020multilogue,Lv_2021_CVPR,bedi2021multi}\end{tabular} \\ \toprule
\begin{tabular}[c]{@{}c@{}}Speech enhancement \\ and separation\end{tabular}        & \multirow{2}{*}{\begin{tabular}[c]{@{}c@{}}Unknown sources (speaker or objects) are \\ hard to be split purely based on audio.\end{tabular}} & \begin{tabular}[c]{@{}c@{}}Speakers in visual are often isolated.\\ Visual has consistent speech content.\end{tabular}             & \begin{tabular}[c]{@{}c@{}}\cite{hou2018audio,ideli2019visually,michelsanti2019training,afouras2018conversation}\\ \cite{afouras2019my,sadeghi2020robust,lee2021looking,kang2022impact}\end{tabular} \\ \cline{3-4} 
Object sound separation                                                             &                                                                                                                                            & \begin{tabular}[c]{@{}c@{}}Objects in visual are often isolated.\\ Motion and produced sound  are consistent.\end{tabular}         & \begin{tabular}[c]{@{}c@{}}\cite{gao2018learning,zhao2018sound,rouditchenko2019self,zhao2019sound}\\ \cite{xu2019recursive,gao2019co,chatterjee2021visual,tian2021cyclic}\end{tabular}               \\ \cline{2-4} 
\begin{tabular}[c]{@{}c@{}}Face super-resolution \\ and reconstruction\end{tabular} & \begin{tabular}[c]{@{}c@{}}Low-resolution or fake images loss \\ personality identities.\end{tabular}                                      & \begin{tabular}[c]{@{}c@{}}Face-voice have consistent personality identities.\\ Voice can indicate facial attributes.\end{tabular} & \cite{meishvili2020learning,kong2021appearance}                                                                                                                                                      \\ \bottomrule
\end{tabular}}
\end{table*}

\subsection{Uni-modal Enhancement}
The consistency between audio-visual modalities not only provides the foundation for the fusion of multiple modalities but also makes it possible to enhance the uni-modal signal. For example, the isolated visual information of multiple speakers can assist in the separation of mixed speech, and audio information can also reflect the covered or missing facial information. These phenomena have inspired researchers to use the information of one modality as an aid to enhance or denoise the content of another.

\textbf{Speech Enhancement and Separation.} Speech enhancement and separation are two tightly related tasks, which aim to recover clean target speech from a mixture of sound (\emph{e.g.,} overlapped speech of multiple speakers or speech with background noise). In the beginning, these tasks take the audio signal as the input only~\cite{vincent2018audio}, making them sensitive to the acoustic noise. As stated above, human speech perception relies on both sight and hearing. The motion of lips, tongue, as well as facial expressions, also reveal related semantics information. Meanwhile, visual information is often unaffected by acoustic noise and multiple speakers are visually isolated. Hence, the visual modality is brought into speech enhancement and separation tasks for generating high-quality audio. In the infancy of these studies, the knowledge-based methods and classical statistical approaches are proposed, including non-negative matrix factorization~\cite{schmidt2006single}, mutual information~\cite{fisher2000learning}, HMMs~\cite{hershey2001audio} and visually-derived Wiener filter~\cite{girin2001audio,almajai2010visually}. Later, the deep audio-visual models have demonstrated impressive performance on reconstructing clean speech signal with different forms (\emph{e.g.,} waveform~\cite{ideli2019visually} and spectrogram~\cite{hou2018audio,michelsanti2019training}) with the assistance of lip movement~\cite{gabbay2018visual,afouras2018conversation}, face clips~\cite{gabbay2018seeing,ochiai2019multimodal,lee2021looking} and even static face image~\cite{chung2020facefilter}. Overall, the introducing of visual information strengthens the ability of speech enhancement and separation models by providing auxiliary information.

\textbf{Object Sound Separation.}
Similar to the face-voice association in the speech situation, scenarios where simultaneously exist multiple sounding objects, are common in our daily life. Object sound separation aims to isolate the sound produced by the specific object from the mixture. Since the visual objects in the same semantic category usually produce similar sound pattern, researchers proposed to introduce the static visual appearance of object to split its audio signal from the mixed ones of different categories objects~\cite{gao2018learning,zhao2018sound}. 
However, these methods cannot well handle the scenarios that sounding objects are with the same category, since the objects appearance are visually similar. 
Thus, the motion information, like the trajectory cues~\cite{zhao2019sound} and keypoint-based structural representation~\cite{gan2020music}, is introduced, by further considering the temporal and spatial consistency between audio and visual modalities.
To better build the audio-visual mapping, most methods above employ the strategy that mixes existing audio-visual segments, then learns to reconstruct each of them, to train the model. Although they have achieved significant results, the acoustic properties as well as visual environment information of real multi-source videos are practically neglected in the synthetic ones. 
Hence, Gao \emph{et al.}~\cite{gao2019co} proposed the co-separation training paradigm that learns object-level audio-visual correlation in realistic multi-source videos.
Recently, considering the same object may produce diverse sounds, Chatterjee \emph{et al.}~\cite{chatterjee2021visual} formulated the visual scene as a graph, then modelled the characteristics of different interactions for the same object. In general, sound separation has gradually evolved in many aspects, including the difficulty degree of separation and training paradigm.

\textbf{Face Super-resolution and Reconstruction.} As stated in the speaker recognition task, voice and face own consistent personality identities. Therefore, it is achievable to estimate important facial attributes such as gender, age as well as ethnicity based on someone's voice. Hence, audio can be the assistance in image processing when the visual information is blurred or missing, like face super-resolution and reconstruction. Face super-resolution is the task which aims to recover details of a limited resolution image. Meishvili \emph{et al.}~\cite{meishvili2020learning} introduced the audio signal that carries facial attributes to recover a plausible face with higher resolution, where the facial-aware audio embedding is fused with the visual ones.
In addition, the deepfake attack, which aims to automatically manipulate the face in a video, has been with a broader research interest recently.
The face-voice consistency in personality identity provides a solution for defending against this attack. Kong \emph{et al.}~\cite{kong2021appearance} leveraged the joint information of face and voice to reconstruct the authentic face given the fake face as well as the voice signal. Overall, the correlation between voice and face makes it promising to enhance visual information with the aid of audio.

\subsection{Discussion}
Based on the above review and analysis of audio-visual boosting tasks, we conclude the problem of uni-modal input as well as the motivation of audio-visual input in~\autoref{tab:boost}. 
The tasks in this section aim to boost the performance of existing uni-modal tasks via combining audio-visual information, which is supported by the audio-visual consistency across semantic, spatial and temporal. Although these tasks target at different aspects of audio-visual recognition and enhancement, they are not isolated from each other and can be connected to build models with broader audio-visual application scenes.
For example, the speech separation task can be combined with the speaker diarization task to handle the overlapped speech scenarios. In addition, consistency is essentially the core of audio-visual learning, but modality properties should not be neglected. 
Recent studies found that the performance of the audio-visual model is not always superior to the uni-modal one, especially in the recognition task. The reason is that the learning characteristics of different modalities were regretfully ignored in the training~\cite{wang2020makes,xiao2020audiovisual,peng2022balanced}. To fully exploit audio-visual information for boosting single modality learning, these modal-specific properties should also be considered.

\begin{figure*}[t]
    \centering
    \includegraphics[width=\linewidth]{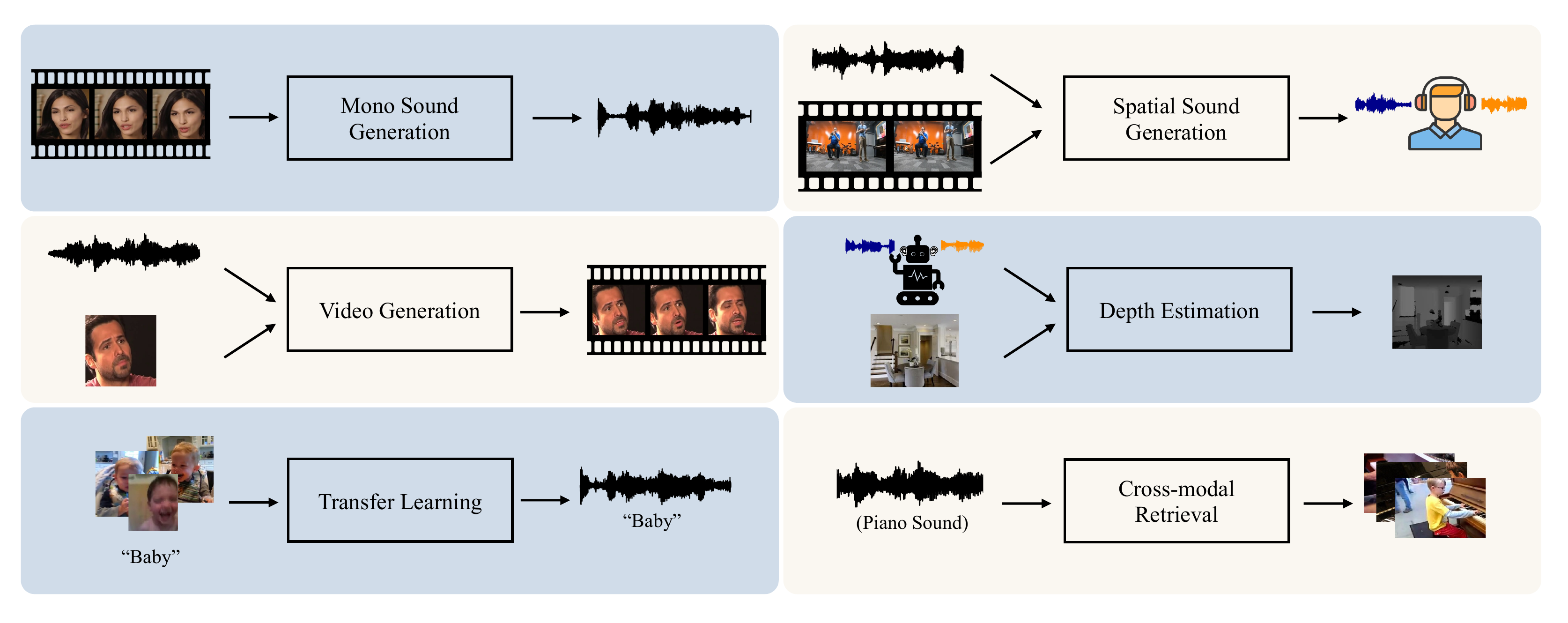}
    \caption{\textbf{Illustration of the cross-modal perception tasks.} The cross-modal perception tasks include cross-modal generation, audio-visual transfer learning and cross-modal retrieval. The cross-modal generation tasks often emphasize audio-visual spatial-temporal consistency, like the changes in facial region or received sound, besides the semantic-level consistency. The transfer learning, as well as cross-modal retrieval tasks, aim to transfer or search the semantic-related information.}
    \label{fig:cross}
\end{figure*}

\section{Cross-modal Perception}
\label{sec:cross}
The cross-modal plasticity phenomenon in human cognition as well as audio-visual consistency prompt the studies of cross-modal perception, which aims to learn and build association between audio and visual modalities, enabling cross-modal generation, transfer and retrieval. The outlines of these cross-modal perception tasks are provided in~\autoref{fig:cross}.

\subsection{Cross-modal Generation}
Humans have the ability to predict information of one modality with the guidance of others. For example, even only having the visual information of lip motion, we can infer what the person is speaking. The rich audio-visual consistency makes it possible for machines to mimic human-like cross-modal generation ability.

\textbf{Mono Sound Generation.} The correlation between motion and produced sound provides the possibility to reconstruct the audio signal based on silent videos. The scenes of mono sound generation cover speech reconstruction, music generation as well as natural sound generation. Under the speech scenarios, the speech content can be revealed by visual information, including the movement of lips and tongue. This fact inspires the speech reconstruction task which aims to generate the audio signal from silent videos.
Early studies of this task proposed to estimate the spectral envelope from the hand-crafted visual feature, then the spectral envelope is used to synthetic corresponding audio signal~\cite{milner2015reconstructing,le2017generating}. Since it is time-consuming and hard to construct suitable hand-crafted visual features, the subsequent works learnt to extract visual features from raw data~\cite{ephrat2017vid2speech,ephrat2017improved,akbari2018lip2audspec}. 
To avoid the burden of learning audio-related intermediate representation, the \emph{Generative Adversarial Networks} (GAN) based model is proposed to directly synthesize audio waveform from silent videos~\cite{vougioukas2019video,mira2022end}. 
Moreover, considering video of one single view is hard to capture complete information, multi-view videos (\emph{e.g.,} frontal and profile views) are used to improve generation quality~\cite{kumar2018harnessing,uttam2019hush}. 

Musical scene is another typical audio-visual scenario. To generate high-quality music, intermediate representations, like musical instrument digital interface, are utilized to bridge the visual changes of playing instruments and generated audio.
Early studies of music generation focused on relatively simple musical scenes, especially the piano case, using traditional computer vision techniques to capture the changes of instrument~\cite{akbari2015real}. More recently, deep learning-based models, including Variational Autoencoder and GANs, have emerged to capture the playing motion in terms of body keypoints~\cite{gan2020foley,su2020multi} and movements~\cite{kurmi2021collaborative} in general instrument scenes.
The above methods aim to produce determined music based on strict audio-visual correspondence, while Di \emph{et al.}~\cite{di2021video} leveraged the rhythmic relations between video and music at the semantic level to generate non-determined background music, matching the given video.

Further, the sound generation of natural scenes is more challenging since the audio-visual correspondence in realistic scenes is hard to be captured, compared with the limited speech or musical scenes. Owens \emph{et al.}~\cite{owens2016visually} pioneered the natural sound generation task via collecting the Greatest Hits dataset and presented a model that generates sound based on visual changes of hitting or scratching objects. Subsequently, Zhou \emph{et al.}~\cite{zhou2018visual} employed the SampleRNN model to generate raw waveform samples given the visual appearance and optical flow, covering ambient sounds and sounds of animals or peoples. Chen \emph{et al.}~\cite{chen2020generating} took the object-irrelevant background sound into consideration, achieving high-quality results. Recently, Iashin \emph{et al.}~\cite{iashin2021taming} presented an efficient visual-driven generation method via a codebook representation that is trained by a variant of GANs. Generally, mono sound generation has been a flourishing field, covering multiple audio-visual scenarios.

\textbf{Spatial Sound Generation.} Humans' audition system can determine the location of hearing sounds, which is called the binaural effect. The recorded binaural audio is then expected to recover the stereo sound sensation. However, recording such a spatial sound requires special equipment that is not easily accessible. Hence, the audio-visual spatial consistency inspires researchers to generate spatial sound with the guidance of visual information. Li \emph{et al.}~\cite{li2018scene} proposed to generate the stereo sound of a specific room via integrating synthesised early reverberation and measured late reverberation tail, while Morgado \emph{et al.}~\cite{morgado2018self} adopted an end-to-end deep neural framework for ambisonic sound generation using $360^{\circ}$ videos. However, $360^{\circ}$ videos are limited in amounts and scenarios, making these methods cannot generalize to other environments well. 
Subsequently, some studies are proposed to use the \emph{Normal Field of View} (NFOV) videos for reconstructing binaural audio from mono sound, driven by the visual location of the sounding source~\cite{gao20192,lu2019self,parida2022beyond}.
These methods are built in a data-driven way, relying on the ground truth stereo sound as the supervision. Although they achieve promising performance, the supervised manner narrows their application to the specific scenes. Thus, Zhou \emph{et al.}~\cite{zhou2020sep} formulated the spatial sound generation task into an extreme case of sound separation, combining stereo sound and mono sound as training data to improve model performance and generalization. To further alleviate the dependence on the insufficient and expensive stereo sound, Xu \emph{et al.}~\cite{xu2021visually} utilized the head-related impulse response and spherical harmonic decomposition to build pseudo visual-stereo pairs from mono sound data. Instead, Lin \emph{et al.}~\cite{lin2021exploiting} exploited the association between each audio component and the spatial regions of interest to explore the audio-visual spatial consistency. As stated, spatial sound generation methods gradually alleviate reliance on ground truth stereo videos, expanding their application scenarios.

\textbf{Video Generation.} Besides video-to-sound generation, sound-to-video generation is also a research field with much attention, which contains speech-based cases and music-based cases. The speech-based generation mainly includes talking face/head generation and gesture generation. 
The former task aims to produce determined videos based on the input speech, and the latter focuses on predicting plausible content-related gestures. 
The early talking face generation methods are speaker-dependent and rely on a large video corpus of the target person~\cite{jamaludin2019you,suwajanakorn2017synthesizing}. For example, Suwajanakorn \emph{et al.}~\cite{suwajanakorn2017synthesizing} proposed to generate the talking image via retrieving the appropriate lip region in the video footage of the target person. Subsequently, speaker-independent face generation methods in an audio-driven manner are presented~\cite{chen2018lip,vougioukas2020realistic}. These methods only focus on the generation of face or lip region, while when a person is speaking, other parts (\emph{e.g.,} facial expression and head pose) besides lips are also variable. Hence, researchers introduced structural information, such as landmark~\cite{zakharov2019few,zhou2020makelttalk,zhang2020freenet} and 3D model~\cite{thies2020neural,zhou2020rotate,li2021write}, to model the correlation between speech and more general facial parts.
Zhou \emph{et al.}~\cite{zhou2021pose} further proposed an implicit low-dimension pose code to generate the pose-controllable talking face without the assistance of structural information. Such strategy could avoid the performance degradation caused by inaccuracy estimated structural representation. 
Recently, the emotion information is disentangled from the speech to drive more fine-grained talking face generation~\cite{ji2021audio}.

Talking alongside gestures is common for humans, which is helpful to emphasize specific information during communication. The alignment between gesture and speech is implicit and ambiguous, bringing difficulty to its modelling. In the early stage, the gesture generation methods are mostly rule-based, resulting in the produced gesture being limited to a selected discrete set~\cite{cassell2004beat,huang2012robot}. With the advance of deep learning, more methods begin to use the data-driven scheme to fully model the speech-gesture alignment pattern of different speakers, utilizing 2D~\cite{ginosar2019learning,liang2022seeg} or 3D pose model~\cite{kucherenko2019analyzing,hasegawa2018evaluation}. The early methods formulated this task as a classification problem~\cite{ahuja2019react}, while more recent works consider it as the regression problem to generate continuous gesture~\cite{alexanderson2020style,kucherenko2020gesticulator}. Since each speaker often owns a specific speech-gesture alignment style, Ginosar \emph{et al.}~\cite{ginosar2019learning} proposed to model this speaker-specific style during generation. Ahuja \emph{et al.}~\cite{ahuja2020style} further transferred the learnt style of one speaker to another via disentangling the style and content of gestures. Recently, Liang \emph{et al.}~\cite{liang2022seeg} decoupled speech information into semantic-relevant cues and semantic-irrelevant cues to explicitly learn and produce semantic-aware gestures.

\begin{table*}[!t]
\renewcommand\arraystretch{1.3}
\centering
\caption{A summary of cross-modal perception tasks.}
\label{tab:cross}
\setlength{\tabcolsep}{0.5mm}{
\begin{tabular}{cccc}
\toprule
Task                           & Motivation                                                                                                                      & Purpose                                                                                                    & Representative methods                                                                                                                                                                                          \\ \toprule
Mono sound generation          & Motion and produced sound are consistent.                                                                                       & \begin{tabular}[c]{@{}c@{}}Reconstruct audio signal \\ based on silent videos.\end{tabular}                & \begin{tabular}[c]{@{}c@{}}\cite{owens2016visually,akbari2018lip2audspec,vougioukas2019video}\\ \cite{vougioukas2019video,su2020audeo,gan2020foley,kurmi2021collaborative}\end{tabular}                        \\ \cline{2-4} 
Spatial sound generation       & \begin{tabular}[c]{@{}c@{}}Visual scenes can reveal\\ spatial information of sound.\end{tabular}                                & \begin{tabular}[c]{@{}c@{}}Generate spatial sound \\ with the guidance of visual.\end{tabular}             & \begin{tabular}[c]{@{}c@{}}\cite{gao20192,lu2019self,zhou2020sep}\\ \cite{parida2022beyond,xu2021visually,lin2021exploiting}\end{tabular}                                                                       \\ \cline{2-4} 
Video generation              & \begin{tabular}[c]{@{}c@{}}Visual changes of sound source  \\ are related to the audio.\end{tabular}  & \begin{tabular}[c]{@{}c@{}}Generate content-relate motion \\ based on audio.\end{tabular}                  & \begin{tabular}[c]{@{}c@{}}\cite{suwajanakorn2017synthesizing,zakharov2019few,lee2019dancing,zhou2020makelttalk}\\ \cite{zhou2021pose,alexanderson2020style,li2021ai,liang2022seeg}\end{tabular} \\ \cline{2-4} 
Depth estimation               & \begin{tabular}[c]{@{}c@{}}Binaural sound can indicate\\ spatial information of visual scenes.\end{tabular}                     & \begin{tabular}[c]{@{}c@{}}Estimate depth information \\ with the guidance of binaural sound.\end{tabular} & \cite{christensen2020batvision,gao2020visualechoes,parida2021beyond,irie2022co}                                                                                                                                 \\ \toprule
Audio-visual transfer learning & \multirow{2}{*}{\begin{tabular}[c]{@{}c@{}}Audio and visual modalities\\ are related in semantics.\end{tabular}} & \begin{tabular}[c]{@{}c@{}}Strengthen modal capacity via \\ transferring the knowledge from others.\end{tabular}     & \begin{tabular}[c]{@{}c@{}}\cite{aytar2016soundnet,owens2016ambient,gan2019self}\\ \cite{yin2021enhanced,xue2021multimodal,chen2021distilling}\end{tabular}                                                     \\ \cline{3-4} 
Cross-modal retrieval          &                                                                                                                  & \begin{tabular}[c]{@{}c@{}}Retrieve content-related samples \\ in the cross-modal scenario.\end{tabular}      & \cite{fan2011example,lee2013music,hong2017content,li2017image2song,suris2018cross}                                                                                                                                                 \\ \bottomrule
\end{tabular}}
\end{table*}

Dancing to music is another sound-to-video generation scenario. The exploration of the alignment between dance and music characteristics (\emph{e.g.,} beat, rhythm and tempo) has attracted much attention for a long time. The early works of dance generation tackle this task as a retrieval problem that mainly consider the motion-music similarity, limiting the diversity of dance~\cite{fan2011example,ofli2011learn2dance,lee2013music}. Later, models, like \emph{Long Short-Term Memory} (LSTM), were widely used to predict motion and pose given a music~\cite{tang2018dance}. Recently, the dance generation task has been formulated from a generative perspective and achieved promising performance~\cite{lee2019dancing,huang2020dance,li2021ai}. For example, Lee \emph{et al.}~\cite{lee2019dancing} adopted the decomposition-to-composition learning framework to temporally align and synthesize dance with accompanying music. Further, Huang \emph{et al.}~\cite{huang2022genre} took the music genre into consideration, beyond the alignment in terms of beat and rhythm. 
Besides dancing to music, Shlizerman \emph{et al.}~\cite{shlizerman2018audio} focused on the scenario of playing instruments. They proposed to generate plausible motion of playing given music. 
Apart from these video generation tasks spanning various scenarios, several related methods are also proposed to manipulate the image content according to audio information~\cite{lee2022sound,li2022learning_image}. In a nutshell, the sound-to-video generation has produced many extraordinary works, which have the potential to facilitate a wide range of applications in practice.

\textbf{Depth Estimation.} Apart from the spatial sound generation based on visual input, it is feasible to use spatial sound to estimate the spatial information of visual scene, especially in low-light or no-light conditions. The spatial consistency between audio and visual modalities makes this task achievable. 
Christensen \emph{et al.}~\cite{christensen2020batvision} proposed the BatVision system that predicts depth maps purely based on binaural echo information. Meanwhile, 
Gao \emph{et al.}~\cite{gao2020visualechoes} offered a new training fashion that interacts with the environment in an audio-visual way. They fused the monocular image features and binaural echo features to improve the depth estimation quality. 
Beyond simply fusing audio-visual features, Parida \emph{et al.}~\cite{parida2021beyond} explicitly took the material properties of various objects within a scene into consideration, significantly enhancing the depth estimation performance. The above methods tend to use a single audio feature, typically spectrogram, and other audio features are seldom been paid attention to. Hence, Irie \emph{et al.}~\cite{irie2022co} introduced angular spectrum, which is useful in geometric prediction tasks, achieving impressive estimation results. Overall, audio-visual depth estimation is a developing field that promises to expand into more realistic scenarios of application.

\subsection{Audio-visual Transfer Learning}
The audio-visual consistency in semantics suggests that the learning of one modality is expected to be aided by the semantic knowledge of the other one. This is the aim of the audio-visual transfer learning task. 
To avoid the expensive and time-consuming labelling process, Aytar \emph{et al.}~\cite{aytar2016soundnet} designed a teacher-student network to train the student audio model with a pre-trained vision teacher, by resorting to large-scale and economically acquired unlabelled videos.
Correspondingly, Owens \emph{et al.}~\cite{owens2016ambient} proposed to use the ambient sound as a supervisory signal to learn visual representations. Afterwards, Gan \emph{et al.}~\cite{gan2019self} also trained the stereo sound student model by transferring the knowledge of the visual teacher for the vehicle tracking task. Their model can independently localize the object purely based on stereo sound during the test. 
The above strategy only uses the RGB visual teacher during training, which limits their performance, since RGB modality is vulnerable to the variants of factors like weather and illumination. Hence,
Valverde \emph{et al.}~\cite{valverde2021there} combined multiple visual teacher networks, including RGB, depth, and thermal, to fully utilize the complementary visual cues for improving the robustness of audio student network. 
Similarly, Yin \emph{et al.}~\cite{yin2021enhanced} designed a robust audio-visual teacher network to fuse complementary information of multiple modalities, facilitating the learning of the visual student network. Instead, Zhang \emph{et al.}~\cite{zhang2021knowledge} distilled the knowledge of audio-visual teachers at three different levels: label-level, embedding-level and distribution level. To sum up, the above methods strengthen the distillation performance by increasing teacher numbers or distillation levels.
In contrast, Xue \emph{et al.}~\cite{xue2021multimodal} used a uni-modal teacher to train a multi-modal student. They found that the student could correct inaccurate predictions and generalize better than its uni-modal teacher. 
However, the employed videos for transfer are often ``in the wild'', thus visual can be unaligned with the accompanying audio (\emph{e.g.,} audio is background music), which brings noise for transfer learning. Chen \emph{et al.}~\cite{chen2021distilling} considered this scenario and proposed to capture the task-related semantics via compositional contrastive learning for robust transfer.
Overall, existing methods strengthen the transfer quality in terms of number of teachers, transfer levels as well as transfer robustness.

\subsection{Cross-modal Retrieval}
Cross-modal retrieval in the audio-visual scenario is another typical cross-modal perception task. Its purpose is to retrieve data of one modality based on the query of another, which has been a rapid development research area in these years~\cite{hong2017content,suris2018cross,zeng2020deep,chen2020deep,asano2020labelling}. Essentially, semantic associations build the bridge between modalities. Even though, the heterogeneous forms of audio-visual modalities make it necessary to map their representations into the same space, where the distance of data pairs reflects the semantic similarity. Variants of \emph{Canonical Correlation Analysis} (CCA) are broadly utilized in the cross-modal retrieval task. They aim to find transformations of two modalities via maximizing the pairwise correlation including Kernel-CCA~\cite{lai2000kernel}, Deep-CCA~\cite{andrew2013deep} and Cluster-CCA~\cite{rasiwasia2014cluster}. Besides CCA-based methods, some researchers introduced other constraints to train the audio-visual model with a joint embedding space. Sur{\'\i}s \emph{et al.}~\cite{suris2018cross} utilized both the cosine similarity loss and classification loss to project the uni-modal features into the common feature space.
Hong \emph{et al.}~\cite{hong2017content} used the ranking loss between modalities to improve the semantic similarity of video-music pairs. 
Instead, Hu \emph{et al.}~\cite{li2017image2song} proposed to build the correlation between image and music, with the assistance of text.
These retrieval methods has been applied many cross-modal perception tasks, including music recommendation given video~\cite{hong2017content,zeng2018audio} and dance generation~\cite{fan2011example,lee2013music}.

\subsection{Discussion}
The motivation and purpose of the cross-modal perception tasks are concluded in~\autoref{tab:cross}. 
The core of these tasks is to perceive one modality with the input of another, relying on audio-visual consistency. Cross-modal retrieval aims to retrieve content-related samples but of another modality, since the audio-visual corresponding pairs own similarities in semantics. Furthermore, the audio-visual transfer learning task transfers the learnt knowledge from one modality to others, enhancing the modal capacity. Beyond the semantic-level perception, the cross-modal generation tasks often require the audio-visual spatial-temporal consistency to model the fine-grained cross-modal correlation. For example, changes in facial regions, especially the mouth area, should be accurately corresponding to the speech in the talking face generation task. These individual generation tasks, like talking face generation, dance generation and gesture generation, usually focus on the single aspect of humans and thus are promising to further be integrated for applications such as virtual humans synthesis. What's more, the audio and visual, as two different modalities, also contain modality-specific information, which is hard to be directly predicted using cross-modal correlation. How to both consider the modality-shared and modality-specific information in the cross-modal perception process, is still an open question.

\begin{figure*}[t]
    \centering
    \includegraphics[width=\linewidth]{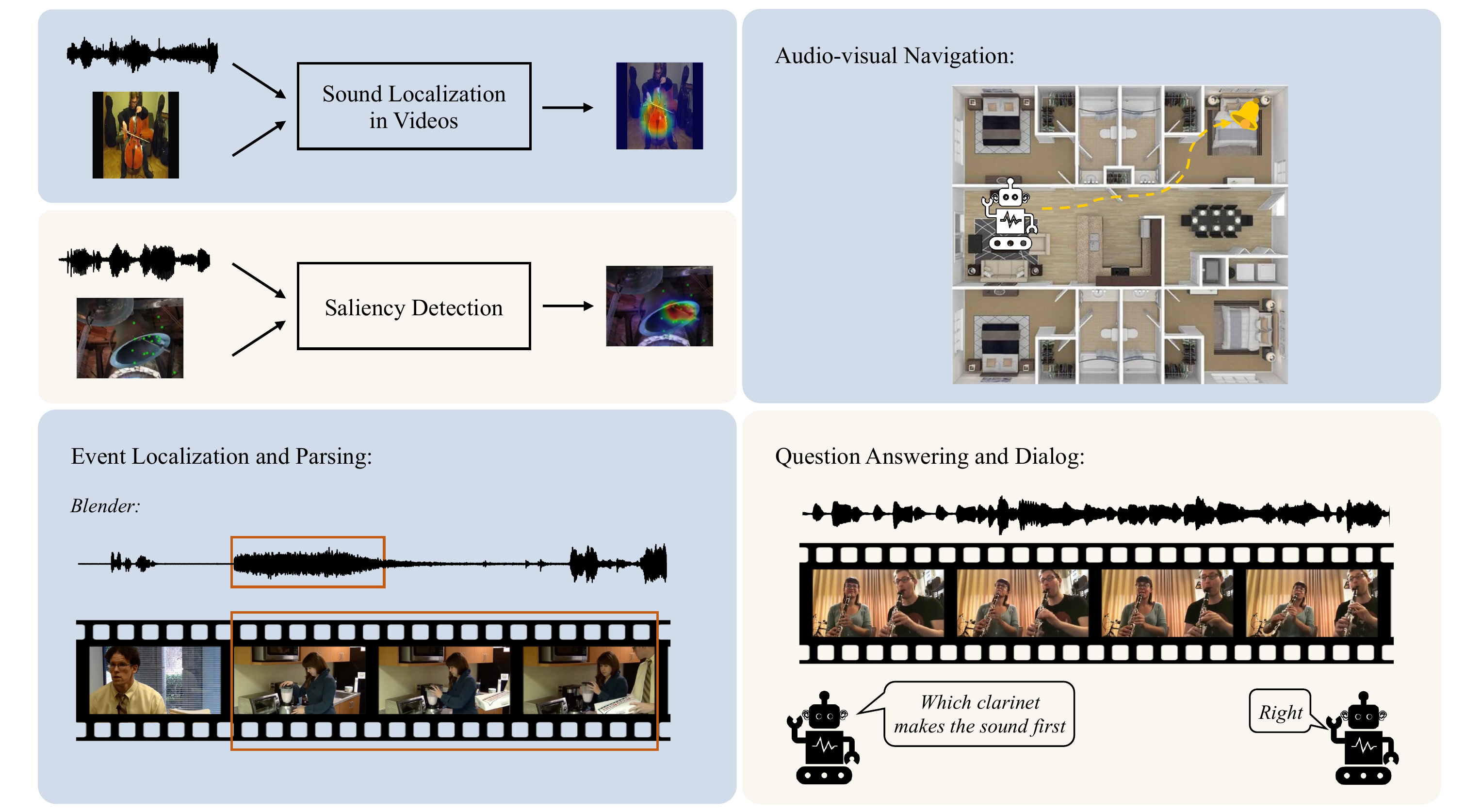}
    \vspace{-2em}
    \caption{\textbf{Illustration of the audio-visual collaboration tasks.} Besides audio-visual representation learning, the audio-visual collaboration tasks include audio-visual localization, audio-visual event localization and parsing, as well as audio-visual question answering and dialog. These tasks pay attention to the perception of audio-visual components or reasoning in audio-visual scenes, collaboratively utilizing the audio-visual modalities to decouple and understand the scenes.}
    \vspace{-1em}
    \label{fig:both}
\end{figure*}

\section{Audio-visual Collaboration}
\label{sec:both}
The audio-visual information is integrated in a collaborative manner, which plays an important role in human scene understanding ability. Hence, it is necessary for the machine to pursue human-like perception via exploring the audio-visual collaboration, beyond just fusing or predicting multi-modal information. 
For this goal, new challenges spanning from simple scene understanding to audio-visual components interaction as well as reasoning are introduced. \autoref{fig:both} offers the sketches of these audio-visual collaboration tasks.

\subsection{Audio-visual Representation Learning}
How to effectively extract representation from heterogeneous audio-visual modalities without human annotations, is an important topic. This is because that high-quality representation can contribute to a variety of downstream tasks~\cite{baltruvsaitis2018multimodal}. For audio-visual data, the consistencies across semantic, spatial as well as temporal between them, naturally provide supervision for representation learning. Not always but often, the visual and accompanying sound is synced in temporal and semantically consistent. Meanwhile, the spatial location in stereo scenarios is also consistent. These consistencies provide ideal supervision for audio-visual representation learning. In the early stage, de Sa~\emph{et al.}~\cite{de1994learning} proposed to minimize the audio-visual disagreement to learn from unlabelled data.
Arandjelovi\'{c} \emph{et al.}~\cite{arandjelovic2017look} employed the semantic consistency and proposed the \emph{Audio-Visual Correspondence} (AVC) learning framework whose goal is to determine whether the audio input and visual input are from the same video or not. Korbar \emph{et al.}~\cite{korbar2018cooperative} further proposed the Audio-Visual Temporal Synchronization, which aims to decide whether a given audio input and a visual input are either in-sync or out-of-sync, considering the more strict audio-visual consistency in temporal. Besides, Morgado \emph{et al.}~\cite{morgado2020learning} leveraged the spatial consistency between $360^{\circ}$ video and spatial audio as supervision to learn audio-visual representation, revealing advantages on a variety of downstream tasks. 
The above primary methods are simple yet effective, but suffer from the data noise when training with ``in the wild'' videos~\cite{morgado2021robust}. 
Some recent methods further view the false negative pairs and false positive pairs as the noise~\cite{morgado2021audio,morgado2021robust}. They proposed to estimate the quality of audio-visual pairs, then optimize with a weighted contrastive learning loss, to alleviate the impact of false pairs.
Further, to better learn representation in the scenes with multiple sound sources, Deep Multimodal Clustering model is proposed to first disentangle individual sources of each modality, then associate them across modalities~\cite{hu2019deep}.
The learnt representation by these methods shows impressive results in multiple downstream tasks, such as acoustic scene classification and image classification, sometimes even surpassing human performance~\cite{korbar2018cooperative,hu2019deep}.

The above methods are often performed in a pairwise fashion, which usually lack the modelling of the sample distribution. Therefore, deep clustering techniques are introduced. Alwassel \emph{et al.}~\cite{alwassel2020self} first proposed the \emph{Cross-Modal Deep Clustering} (XDC) method to learn representation in a self-supervised manner. They trained the audio and visual model alternatively with the cross-modal pseudo label. Compared with the XDC method with single modality clusters, Asnao \emph{et al.}~\cite{asano2020labelling} regarded multiple modalities as different data augmentation forms, then learnt audio-visual clusters to annotate the unlabelled videos. Furthermore, Chen \emph{et al.}~\cite{Chen_2021_ICCV} combines the pair-wise contrastive learning scheme and clustering methods to fully capture semantic consistency. These deep audio-visual clustering methods provide another path to learning representation in a self-supervised manner.

In recent years, with the increase in computation sources and accessible data, pre-training models that use pretext tasks to train on massive data with a large-scale number of parameters have emerged. In the beginning, the pre-training models flourish in the natural language processing field~\cite{radford2018improving,brown2020language}, then inspires the appearance of multi-modal pre-training models. Omni-perception PreTrainer~\cite{liu2021opt} is the first visual-text-audio pre-training model. They designed multi-modal pretext tasks for training at sample-level, token-level and modality-level, covering masked modality modelling, generation and contrastive learning schemes. In the training of VATT~\cite{akbari2021vatt} and AudioCLIP~\cite{guzhov2022audioclip} model, the contrastive learning approach is also utilized to maximize cross-modal similarities. Zellers \emph{et al.}~\cite{zellers2022merlot} proposed a new contrastive masked span learning objective. They trained the model by figuring out which span of audio or text is masked out in the video sequence, besides contrastive learning constrain. 
This pre-training strategy is claimed to enable rapid training as well as powerful generalization ability. The multi-modal pre-training model containing audio-visual modalities is now still in its early stages of development and is promising to greatly advance the generalizability under more scenarios.

\subsection{Audio-visual Localization}
The purpose of audio-visual localization is to highlight the visual region or location of interest by referring to the input audio. These tasks build the cross-modal correlation and highlight the audio-visual components in the scenes.

\textbf{Sound Localization in Videos.} In most cases, the sound can be related to its sounding source in visual as the audio and visual often occur simultaneously. The sound localization task aims to find and locate the sound-related regions in videos for the input audio. Early attempts on this task used shallow probabilistic models~\cite{hershey1999audio,kidron2005pixels} or CCA-related methods~\cite{izadinia2012multimodal} to associate audio and visual signal. 
Recently, deep neural networks bring this task into a new stage. 
AVC framework is a typical training scheme. It learns to maximize the similarity between sound features and the visual embeddings of the same sounding object~\cite{arandjelovic2018objects,owens2018audio,senocak2019learning}.
These methods mostly work well in simple single-source scenarios while are hard to handle the cocktail-party scenario with multiple sounding sources. Thus, Zhao \emph{et al.} ~\cite{zhao2018sound,zhao2019sound} combined the audio-visual source separation objective with localization task, and employed the mix-then-separate framework to generate correspondence between pixels and separated audio signals. Differently, Hu \emph{et al.}~\cite{hu2022mix} formulated the image and sound as graphs and adopted a cycle-consistent random walk strategy for separating as well as localizing mixed sounds. The above methods only locate the sounding area in videos but do not consider the semantics of objects. Hu \emph{et al.}~\cite{hu2021class} employed the clustering method to establish a category-representation object dictionary and conducted class-aware sounding object localization in the cocktail-party scenario.
Considering the above methods often only capture the coarse object contour, Zhou \emph{et al.}~\cite{zhou2022audio} released a segmentation dataset with pixel-level annotation of localization. They introduced the audio-visual segmentation task, whose goal is to conduct pixel-level sounding localization. Overall, the sound localization in video methods ground the sounding regions in visual, achieving fine-grained scene understanding.

\textbf{Audio-visual Saliency Detection.} The saliency detection task aims to model the human attention mechanism. In daily life, our attention can be attracted by salient objects or things. Originally, saliency detection tasks only take visual information as the input, which aims to detect eye-attracting components in images or videos~\cite{chen2017video,chen2018bilevel}. Although humans rely on the visual system to perceive the surroundings to a great extent, other senses, especially hearing, can also affect the perception results. To comprehensively explore the human attention mechanism, it is necessary to integrate audio and visual. The early studies in audio-visual saliency detection often utilized traditional hand-craft techniques to detect audio and visual saliency respectively then fused the uni-modal saliency map~\cite{ruesch2008multimodal,schauerte2011multimodal,min2016fixation}. These methods lack the consideration of inter-modal correlation. Recently, a series of deep-based saliency detection methods have emerged~\cite{tavakoli2019dave,tsiami2020stavis,min2020multimodal,jain2021vinet}. Tavakoli \emph{et al.}~\cite{tavakoli2019dave} designed a two-stream framework and used the concatenation fusion strategy, which is a relatively basic model. Further, STAViS model~\cite{tsiami2020stavis} is proposed to combine audio-visual information at multiple stages. However, the dependence on the human eye-fixation label hinders the development of this field, since it is laborious to collect and annotate massive videos. Therefore, Wang \emph{et al.}~\cite{wang2021semantic} designed a strategy for converting audio-visual semantic category tags into pseudo eye-fixation labels. Then they trained the saliency detection model in a weakly-supervised manner. 
Overall, audio-visual saliency detection is a developing field and more in-depth exploration is expected.

\textbf{Audio-visual Navigation.} Besides localizing objects in recorded videos, localizing and navigating in spatial space is natural for humans. In embodied artificial intelligence field, navigation is a vital topic and visual clues are the main information source~\cite{gupta2017cognitive,gupta2017unifying}. However, for humans, hearing can also provide useful spatial information for our exploration of an unfamiliar environment. Thus, the audio-visual navigation task is introduced, aiming to search and locate the target sounding goal in a complex 3D environment ~\cite{chen2020soundspaces}. 
Chen \emph{et al.}~\cite{chen2020soundspaces} adopted an end-to-end multi-modal deep reinforcement learning method to explore navigation policies with a stream of audio-visual observations. Concurrently, Gan \emph{et al.}~\cite{gan2020look} designed a planner for the action sequence generation, based on the captured spatial memory and observed audio-visual information. 
To further improve the efficiency of navigation, Chen \emph{et al.}~\cite{chen2020learning} proposed to set a series of waypoints in a hierarchical model, then a planner plays the path to each waypoint. 
These prior works assume the sounding goal constantly produces a steady repeating sound.
Differently, Chen \emph{et al.}~\cite{chen2021semantic} introduced the semantic audio-visual navigation task that the produced sound of the sounding goal is semantically consistent with the scene (\emph{e.g.,} water dripping in the bathroom).
The above works often navigate in the environment with a single static sounding goal. Thus, researchers attempted to navigate in more complex scenarios with moving and distracting sounds~\cite{younes2021catch,yu2022sound}. 
Relatively, Majumder \emph{et al.}~\cite{majumder2021move2hear} proposed to train the agent for separating the target source in the 3D environment via moving around based on audio-visual observations. In general, existing audio-visual navigation studies have worked well in the simple simulation 3D environment, but are still a subset of human navigation scenarios, and more realistic cases are under-solved.

\begin{table*}[!t]
\renewcommand\arraystretch{1.3}
\centering
\caption{A summary of audio-visual collaboration tasks.}
\vspace{-1em}
\label{tab:both}
\begin{tabular}{cccc}
\toprule
Task                                                                                   & Motivation                                                                                                  & Purpose                                                                                                      & Representative methods                                                                                                                                                                  \\ \toprule
\begin{tabular}[c]{@{}c@{}}Audio-visual \\ representation learning\end{tabular}        & \begin{tabular}[c]{@{}c@{}}Audio-visual consistency\\ provides natural supervision.\end{tabular}            & \begin{tabular}[c]{@{}c@{}}Learn audio-visual representation\\ without annotations.\end{tabular}               & \begin{tabular}[c]{@{}c@{}}\cite{arandjelovic2017look,korbar2018cooperative,hu2019deep,morgado2020learning}\\ \cite{alwassel2020self,asano2020labelling,liu2021opt,zellers2022merlot}\end{tabular} \\ \toprule
Sound localization in videos                                                           & \begin{tabular}[c]{@{}c@{}}Visual and sound both indicate \\ location of target objects.\end{tabular}       & \begin{tabular}[c]{@{}c@{}}Capture the sound-related\\ region in visual.\end{tabular}                        & \begin{tabular}[c]{@{}c@{}}\cite{arandjelovic2018objects,owens2018audio,senocak2019learning}\\ \cite{zhao2018sound,zhao2019sound,hu2021class,hu2022mix}\end{tabular}                    \\ \cline{2-4} 
\begin{tabular}[c]{@{}c@{}}Audio-visual \\ saliency detection\end{tabular}             & \begin{tabular}[c]{@{}c@{}}Mimic the human attention mechanism\\ in the audio-visual scenes.\end{tabular}   & Capture audio-visual salient pattern.                  & \cite{tavakoli2019dave,tsiami2020stavis,min2020multimodal,jain2021vinet}                                                                                                                \\ \cline{2-4} 
Audio-visual navigation                                                                & \begin{tabular}[c]{@{}c@{}}Imitate the human navigation ability \\ in the audio-visual scenes.\end{tabular} & \begin{tabular}[c]{@{}c@{}}Search and locate the \\ sounding goal in 3D environment.\end{tabular}   & \begin{tabular}[c]{@{}c@{}}\cite{chen2020soundspaces,gan2020look,chen2020learning}\\ \cite{chen2021semantic,majumder2021move2hear,younes2021catch,yu2022sound}\end{tabular}              \\ \toprule
\begin{tabular}[c]{@{}c@{}}Audio-visual event \\ localization and parsing\end{tabular} & \begin{tabular}[c]{@{}c@{}}Events are not always audible \\ and visible in the whole videos.\end{tabular}   & \begin{tabular}[c]{@{}c@{}}Temporally ground the event-relevant \\ segments in each modality.\end{tabular} & \begin{tabular}[c]{@{}c@{}}\cite{tian2018audio,lin2019dual,tian2020unified,duan2021audio}\\ \cite{zhou2021positive,lin2021exploring,Wu_2021_CVPR,pasi2022investigating}\end{tabular}    \\ \toprule
\begin{tabular}[c]{@{}c@{}}Audio-visual question \\ answering and dialog\end{tabular}  & \begin{tabular}[c]{@{}c@{}}Model human audio-visual \\ scene-aware reasoning ability.\end{tabular}          & \begin{tabular}[c]{@{}c@{}}Converse or answer questions\\ based on the audio-visual scenes.\end{tabular}     & \begin{tabular}[c]{@{}c@{}}\cite{alamri2019audio,hori2019joint,schwartz2019simple}\\ \cite{geng2021dynamic,yun2021pano,li2022learning,shah2022audio}\end{tabular}                       \\ \bottomrule
\end{tabular}
\end{table*}

\subsection{Audio-visual Event Localization and Parsing}
In most audio-visual tasks, like audio-visual correspondence, the audio and visual context is assumed to be matched over the whole video. But videos, especially unconstrained videos, often contain temporally irrelevant audio-visual segments. For example, only some segments of videos with the action ``playing the basketball'' are both audible and visible for this action, since the camera can move to the audience during the shoot. Hence, Tian \emph{et al.}~\cite{tian2018audio} first introduced the audio-visual event localization task, which aims to temporally demarcate both audible and visible events from a video. They treated this task as a sequence labelling problem. An audio-driven visual attention mechanism is developed to localize sounding objects in visual and a dual multi-modal residual network is designed to integrate audio-visual features. Subsequently, Lin \emph{et al.}~\cite{lin2019dual} used LSTM to solve the event localization in a sequence-to-sequence manner, integrating both global and local audio-visual information. With a similar purpose that capturing global information, Wu \emph{et al.}~\cite{wu2019dual} designed the dual attention matching module to model both the high-level event information as well as local temporal information. Besides, other attention mechanisms are following proposed to explore the inter- and intra-modal correlations~\cite{lin2020audiovisual,duan2021audio}. To filter the interference of irrelevant audio-visual pairs during training, Zhou \emph{et al.}~\cite{zhou2021positive} presented the Positive Sample Propagation method to select audio-visual pairs with positive connections and ignore the negative ones. Further considering the noisy background, Xia \emph{et al.}~\cite{xia2022cross} designed the cross-modal time-level and event-level background suppression mechanism to alleviate the audio-visual inconsistency problem. 

The audio-visual event localization task aims to highlight both audible and visible segments of a video. This strict constraint is hard to well ground the action in the ``in the wild'' video, where the audio and visual information is often not aligned, \emph{e.g.,} the out-of-screen case. Hence, the audio-visual video parsing task, which detects audible, visible and audible-visible events in the temporal dimension, is introduced to have a more fine-grained audio-visual scene understanding~\cite{tian2020unified}. Tian \emph{et al.}~\cite{tian2020unified} formulated the audio-visual parsing as the multi-modal multiple instances learning in a weakly-supervised manner. They adopted a hybrid attention network and multi-modal multiple instance learning pooling method to aggregate and exploit the multi-modal temporal contexts, then discovered and mitigated the noisy label for each modality. To generate more reliable uni-modal localization results, Wu \emph{et al.}~\cite{Wu_2021_CVPR} individually predicted event labels for each modality via exchanging audio or visual track with other unrelated videos. This strategy is based on assumption that the prediction of the synthetic video would still be confident if the uni-modal signals indeed contain information about the target event. In addition, Lin \emph{et al.}~\cite{lin2021exploring} proposed to leverage the event semantics information across videos and the correlation between event categories to better distinguish and locate different events. Later, more audio-visual parsing methods are presented with improvement at feature aggregation or attention module~\cite{pasi2022investigating,wang2022distributed}. Overall, the audio-visual event localization and parsing task ground audio-visual events of each modality temporally, obtaining a more fine-grained perception and understanding of audio-visual scenes.

\subsection{Audio-visual Question Answering and Dialog}
The audio-visual question answering and dialog task aim to conduct cross-modal spatial-temporal reasoning about the audio-visual scenes. The uni-modal question answering tasks have been sufficiently studied, but they are hard to be equipped with more realistic inference ability with partial information of scenes. 
Since sight and hearing are two vital senses in human cognition, the audio-visual question answering task has emerged recently~\cite{yun2021pano,li2022learning}. Yun \emph{et al.}~\cite{yun2021pano} brought up the Pano-AVQA dataset, containing $360^{\circ}$ videos and corresponding question-answer pairs. The Pano-AVQA dataset covers two types of question-answer pairs: spherical spatial relation and audio-visual relation, to better explore the understanding of the panoramic scenes. Li \emph{et al.}~\cite{li2022learning} proposed a large-scale MUSIC-AVQA dataset to facilitate spatial-temporal reasoning under dynamic and long-term audio-visual scenes. The audio-visual scenes in MUSIC-AVQA are mainly musical performances which is a typical multi-modal scenario with abundant audio-visual components and their interaction.
To achieve effective question answering, they first visually grounded the sounding regions, then conducted spatial-temporal reasoning with attention mechanism.

Recently, another active area is audio-visual scene-aware dialog. It aims to train the agent that can converse with humans about a temporally varying audio-visual scene, using natural, conversational language~\cite{alamri2019audio,hori2019joint,hori2019end,shah2022audio}. Compared with question-answering, the scene-aware dialog further considers the context of conversations. Alamri \emph{et al.}~\cite{alamri2019audio} first collected the Audio Visual Scene-Aware Dialog dataset where each sample contains a scene-aware dialog about the video. They provided a simple baseline that leveraged the history dialog and audio-visual input to rank candidate answers. 
Schwartz \emph{et al.}~\cite{schwartz2019simple} adopted an end-to-end training strategy. They used a multi-modal attention mechanism to capture cross-modal interactions, then utilized a LSTM module to generate answers.
To effectively represent then infer on multiple modalities, Geng \emph{et al.}~\cite{geng2021dynamic} presented a spatio-temporal scene graph to capture key audio-visual cues, then used a sequential transformer-based mechanism to highlight the question-aware representations for answer generation.
Overall, the above question answering as well as dialog tasks attempt to explore the audio-visual correlations of events in spatial and temporal, based on the decoupled audio-visual scenes. This is a hopeful direction to better mimic the scene understanding ability of humans.

\subsection{Discussion}
The motivation and purpose of the audio-visual collaboration tasks are summarized in~\autoref{tab:both}.
Tasks of audio-visual boosting and cross-modal perception aim to fuse or predict consistent audio-visual information.
Differently, the audio-visual collaboration tasks focus on the perception of audio-visual components or reasoning in audio-visual scenes. The audio-visual localization task builds cross-modal association to highlight the sounding components, achieving the decoupling of the audio-visual scenes in spatial. After that, the studies of event localization and parsing tasks temporally demarcate the audio events or visual events. Moving forward, question answering and dialog tasks attempt to reason the cross-modal spatial-temporal interaction, targeting at the audio-visual scenes. These tasks gradually decouple and understand the scenes, which are in a phase of rapid development and gain more and more attention.

\begin{table*}[!t]
\renewcommand\arraystretch{1.2}
\scriptsize
\centering
\caption{Some representative datasets in audio-visual learning.}
\label{tab:dataset}
\setlength{\tabcolsep}{1mm}{
\begin{tabular}{ccccccc}
\toprule
Dataset                                                                                                                         & Year                                                       & \# Video & Length  & Data form                                                   & Source             & Representative tasks                                                                                                         \\ \toprule
\begin{tabular}[c]{@{}c@{}}LRW~\cite{chung2016lip}, \\ LRS2~\cite{afouras2018deep} and LRS3~\cite{afouras2018lrs3}\end{tabular} & \begin{tabular}[c]{@{}c@{}}2016,\\ 2018, 2018\end{tabular} & -        & 800h+   & video & in the wild              & \begin{tabular}[c]{@{}c@{}}Speech-related, speaker-related,\\ face generation-related tasks\end{tabular} \\
VoxCeleb~\cite{nagrani2017voxceleb}, VoxCeleb2~\cite{chung2018voxceleb2}                                                        & 2017, 2018                                                 & -        & 2,000h+ & video & YouTube                  & \begin{tabular}[c]{@{}c@{}}Speech-related, speaker-related,\\ face generation-related tasks\end{tabular} \\
AVA-ActiveSpeaker~\cite{roth2020ava}                                                                                            & 2019                                                       & -        & 38.5h   & video                                                          & YouTube                  & Speech-related task, speaker-related task                                                                    \\
Kinetics-400~\cite{kay2017kinetics}                                                                                              & 2017                                                       & 306,245  & 850h+   & video                                                        & YouTube                  & Action recognition                                                                                           \\
EPIC-KITCHENS~\cite{damen2018scaling}                                                                                           & 2018                                                       & 39,594   & 55h     & video                                                         & Recorded videos          & Action recognition                                                                                           \\
CMU-MOSI~\cite{zadeh2016multimodal}                                                                                             & 2016                                                       & 2,199    & 2h+     & video                                                          & YouTube                  & Emotion recognition                                                                                          \\
CMU-MOSEI~\cite{zadeh2018multimodal}                                                                                            & 2018                                                       & 23,453   & 65h+    & video                                                          & YouTube                  & Emotion recognition                                                                                          \\
VGGSound~\cite{chen2020vggsound}                                                                                                & 2020                                                       & 200k+    & 550h+   & video                                                       & YouTube                  & Action recognition, sound localization                                                                       \\
AudioSet~\cite{gemmeke2017audio}                                                                                                & 2017                                                       & 2M+      & 5,800h+ & video                                                        & YouTube                  & Action recognition, sound sepearation                                                                        \\
Greatest Hits~\cite{owens2016visually}                                                                                          & 2016                                                       & 977      & 9h+     & video                                                          & Recorded videos          & Sound generation                                                                                             \\
MUSIC~\cite{zhao2018sound}                                                                                                      & 2018                                                       & 714      & 23h+    & video                                                         & YouTube                  & Sound seperation, sound localization                                                                         \\
FAIR-Play~\cite{gao20192}                                                                                                       & 2019                                                       & 1,871    & 5.2h    & video with binaural sound                                                           & Recorded videos          & Spatial sound generation                                                                                     \\
YT-ALL~\cite{morgado2018self}                                                                                                   & 2018                                                       & 1,146    & 113.1h  & $360^{\circ}$ video                                                          & YouTube                  & Spatial sound generation                                                                                     \\
Replica~\cite{straub2019replica}                                                                                                & 2019                                                       & -        & -       & 3D environment                                                         & 3D simulator & Depth estimation                                                                                             \\
AIST++~\cite{li2021ai}                                                                                                          & 2021                                                       & -        & 5.2h    & 3D video                                                          & Recorded videos          & Dance generation                                                                                             \\
TED~\cite{yoon2019robots}                                                                                                       & 2019                                                       & -        & 52h     & video                                                         & TED talks                & Gesture generation                                                                                           \\
SumMe~\cite{gygli2014creating,tsiami2019behaviorally}                                                                                                  & 2014                                                       & 25       & 1h+     & video with eye-tracking                                                          & User videos              & Saliency detection                                                                                           \\
AVE~\cite{tian2018audio}                                                                                                        & 2018                                                       & 4,143    & 11h+    & video                                                         & YouTube                  & Event localization                                                                                           \\
LLP~\cite{tian2020unified}                                                                                                      & 2020                                                       & 11,849   & 32.9h   & video                                                         & YouTube                  & Event parsing                                                                                                \\
SoundSpaces~\cite{chen2020soundspaces}                                                                                          & 2020                                                       & -        & -       & 3D environment                                                          & 3D simulator           & Audio-visual navigation                                                                                      \\
AVSD~\cite{alamri2019audio}                                                                                                     & 2019                                                       & 11,816   & 98h+    & video with dialog                                                          & Crowd-sourced             & Audio-visual dialog                                                                                          \\
Pano-AVQA~\cite{yun2021pano}                                                                                                    & 2021                                                       & 5.4k     & 7.7h    &   $360^{\circ}$ video with QA                                                        & Video-sharing platforms  & Audio-visual question answering                                                                              \\
MUSIC-AVQA~\cite{li2022learning}                                                                                                & 2022                                                       & 9,288    & 150h+   & video with QA                                                          & YouTube                  & Audio-visual question answering                                                                              \\ \bottomrule
\end{tabular}}
\end{table*}

\section{Dataset}
\label{sec:dataset}
In recent years, the rapid development of mobile terminals as well as the bloom of video-sharing platforms have made it easy to collect massive video-form audio-visual data at a low cost. In~\autoref{tab:dataset}, we list some representative benchmark datasets in audio-visual learning for different tasks. Particularly, some attributes of these datasets are also summarized. According to the table, we can have the following observations:

Firstly, videos of the most recent audio-visual datasets are collected from public video-sharing platforms (\emph{e.g.,} YouTube), which evolve from controlled environment and are much closer to realistic scenes. With the growth of the Internet, on the one hand, more media produce more programs in video form, such as TED, and on the other hand, people are increasingly inclined to share their recorded videos on public platforms, providing easy access to collect datasets. For example, the large-scale speech dataset, LRW~\cite{chung2016lip}, LRS2~\cite{afouras2018deep} and LRS3~\cite{afouras2018lrs3}, consist of BBC news and TED videos. And the Kinetics-400/600/700 datasets~\cite{kay2017kinetics,carreira2018short,carreira2019short} are sourced from user-uploaded YouTube videos. In the future, the audio-visual dataset is expected to cover more realistic scenarios, and more cases, like long-tail and open-set problems, should be further considered. 

Secondly, the scope of audio-visual studies have expanded from conventional video to multiple audio-visual data forms recently, such as $360^{\circ}$ video and 3D environment. Although these new-coming data forms provide more opportunities to explore the audio-visual interaction, they are still limited in scale. This is because the requirement of special equipment makes them not easily accessible in everyday life. 
For instance, the FAIR-Play~\cite{gao20192} dataset for spatial sound generation task only contains $1,871$ videos. More economical ways of acquiring these data are expected. Recently, the 3D simulation environment provides a possible direction, which has been utilized in embodied AI, \emph{e.g.,} SoundSpace for audio-visual navigation~\cite{chen2020soundspaces}. 

Thirdly, label of audio-visual datasets span video-level (\emph{e.g.,} AudioSet~\cite{gemmeke2017audio}), frame-level (\emph{e.g.,} AVA-ActiveSpeaker~\cite{roth2020ava}), event-level (\emph{e.g.,} LLP~\cite{tian2020unified}) and QA-level (\emph{e.g.,} MUSIC-AVQA~\cite{li2022learning}), since the exploration of audio-visual scene is gradually fine-grained. However, the labor-intensive labelling brings obstacles to the creation of datasets, limiting the learning of models. Hence, methods for fine-grained scene understanding tasks (\emph{e.g.,} Audio-visual parsing) tend to train in a semi-supervised or unsupervised manner. It is essential to pay a joint effort to address this challenge in terms of labelling techniques as well as learning schema in the future.

In general, these datasets provide the foundation of audio-visual learning, while there are still some inadequate aspects that the existing dataset is hard to satisfy. Firstly, semantic and spatial-temporal consistency is widely used in audio-visual learning. 
However, the consistency effect on model and task lacks a high-quality benchmark to evaluate. Secondly, the existing large-scale datasets are collected based on information of one modality (\emph{e.g.,} VGGSound~\cite{chen2020vggsound} is audio-based and Kinetics~\cite{kay2017kinetics} is visual-based). 
The AVE~\cite{tian2020unified} and LLP~\cite{tian2020unified} datasets simultaneously consider both audio and visual modality while being limited in their scale. 
Hence, a large-scale audio-visual-aware dataset is expected for the exploration of scene understanding. 
Finally, although many of the existing datasets contain ``in the wild'' data, they are limited in specific scenarios. For example, the object sound separation tasks mainly focus on specific types of videos, like the music scene. More types of realistic scenarios are required in the future to investigate models with better generalization.

\section{Trends and New Perspective}
\label{sec:discussion}
In this section, we first discuss the multiple consistencies between audio-visual modalities, and then provide a new perspective about the scene understanding stage, based on existing audio-visual tasks.

\subsection{Semantic, Spatial and Temporal Consistency}
\label{sec:consistency}
Although the audio-visual modalities have heterogeneous data forms, their inner consistency spans semantic, spatial and temporal, building the basis of audio-visual learning studies. Firstly, the audio and visual signals depict the things of interest from different views. Therefore, the semantics of audio-visual data is considered to be consistent. In audio-visual learning, semantic consistency plays an important role in most tasks. 
For example, this consistency makes it possible to combine audio-visual information for better audio-visual recognition and uni-modal enhancement performance. In addition, the semantic similarity between audio-visual modalities also plays a vital role in cross-modal retrieval and transfer learning. 
Secondly, both audio and visual can help to determine the exact spatial location of the sounding objects. This spatial correspondence also has wide applications. For instance, in the sound localization task, this spatial consistency is utilized to ground the sounding components in visual with the guidance of input audio. 
In stereo cases, spatial consistency is used to estimate depth information with binaural audio or generate spatial sound with visual information. 
Thirdly, the visual object and its produced sound are usually temporally consistent. 
Such consistency is also widely utilized across most audio-visual learning studies, like fusing or predicting multi-modal information in audio-visual recognition or generation tasks.

In practice, these consistencies are not isolated but often co-occur in the audio-visual scene. Hence, they tend to be jointly utilized in relevant tasks. Especially, the combination of temporal and semantic consistencies is the most common case. In simple scenarios, the temporal and semantic consistencies of videos are considered to hold simultaneously. For instance, the audio-visual segments at the same timestamps are thought to be consistent both in temporal and semantics. However, this strong assumption can be failed, such as the ``in the wild'' videos with out-of-screen sound, are abundant. These false positive pairs bring noise during training. Recently, researchers have begun to pay attention to these inconsistent cases between audio and visual information to improve scene understanding quality~\cite{morgado2021robust}. In addition, the combination of semantic and spatial consistencies is also common. For example, the success of sound localization in videos relies on semantic consistency, which helps to explore the spatial location in visual based on input sound. Besides, at the early stage of the audio-visual navigation task, the sounding goal produces a steady repeating sound. Although spatial consistency is satisfied, the semantic content in audio and visual is irrelevant. Later, the semantic consistency of produced sound and location is introduced to improve audio-visual navigation quality.

Overall, the semantic, spatial, as well as temporal consistency of audio-visual modalities, sufficiently support the audio-visual learning studies. The effective analysis and utilization of these consistencies improve the performance of existing audio-visual tasks and contribute to better audio-visual scene understanding.

\begin{figure}[t]
    \centering
    \includegraphics[width=0.8\linewidth]{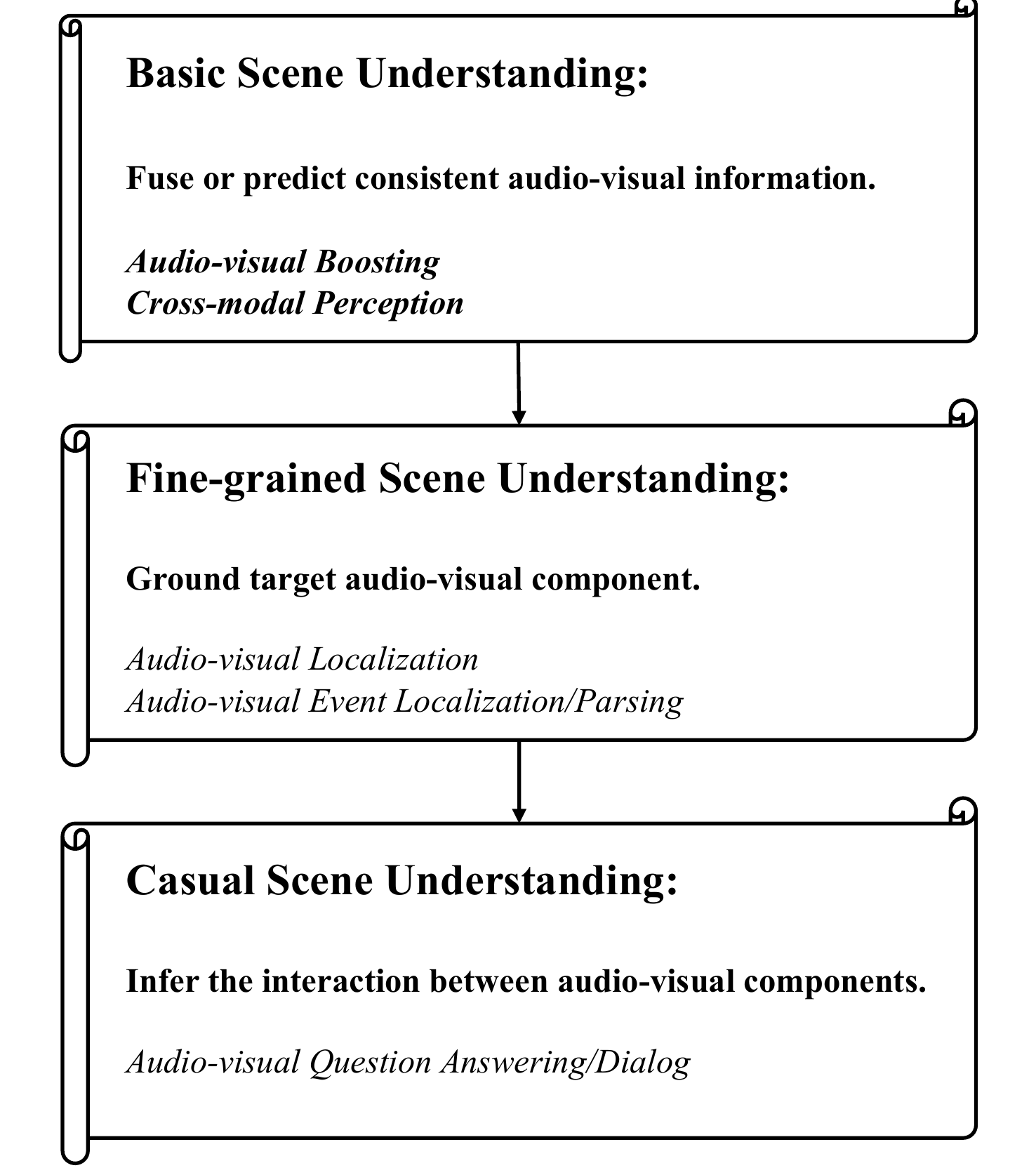}
    \vspace{-1em}
    \caption{Illustration of the new perspective about the audio-visual scene understanding stage.}
    \vspace{-1em}
    \label{fig:new}
\end{figure}

\subsection{New Perspective about Scene Understanding}
In this paper, we conclude the cognition foundation of audio-visual modality and provide several multi-sensory integration phenomenons of humans, based on which, the current audio-visual studies are categorized into three classes: audio-visual boosting, cross-modal perception and audio-visual collaboration. To revisit the current development of audio-visual learning field from a more macro perspective, we further propose a new perspective on audio-visual scene understanding as stated in~\autoref{fig:new}:

\textbf{1) Basic scene understanding.} Tasks of audio-visual boosting and cross-modal perception often focus on fusing or predicting consistent audio-visual information. The core of these tasks is the general understanding of the audio-visual scene (\emph{e.g.,} classifying the action category of input videos.) or predicting based on concurrent cross-modal information (\emph{e.g.,} predicting the speech audio based on the silent videos.). However, videos in natural scenes usually contain various components in audio and visual modality, which could be beyond the scope of these basic scene understanding tasks.

\textbf{2) Fine-grained scene understanding.} As stated above, the audio-visual scenes are often with various modal components.
Hence, several tasks are proposed to further ground the target one. For example, the sound localization in videos task highlights the target sounding objects in visual, and the audio-visual event localization and parsing tasks temporally demarcate audible or visible events of interest. These tasks ground the audio-visual components and decouple the scenes, thus holding a fine-grained understanding compared with the last stage.

\textbf{3) Causal scene understanding.} In the audio-visual scene, humans can not only perceive the surrounding things of interest but also infer the interaction between them. The goal of this stage is closer to the pursuit of human-like perception. Currently, only seldom tasks reach this stage. The audio-visual question answering and dialog tasks are the representative ones. These tasks attempt to explore the correlation of audio-visual components in the videos and conduct spatial-temporal reasoning for achieving better audio-visual scene understanding.

Overall, the current exploration of these three stages is quite imbalanced. From the basic to casual scene understanding, the related studies become not diverse enough and are still in the infancy stage. This indicates several promising future directions in the audio-visual area:

\textbf{Tasks integration for better scene understanding.} Most studies in the audio-visual field are task-oriented. These individual tasks model and learn only specific aspects of the scene. However, the perception of the audio-visual scene is not isolated. For example, the sound localization in videos task emphasizes the sound-related objects in visual while event localization and parsing tasks ground the target event in temporal. These two tasks are promising to be integrated for facilitating the fine-grained understanding of the audio-visual scenes. Hence, the integration of multiple audio-visual learning tasks is worth exploring in the future. 

\textbf{More in-depth causal scene understanding.} The diversity of the studies in casual scene understanding is still limited currently. Existing tasks, including audio-visual question answering and dialog, mostly focus on conversing based on the captured events in videos. More in-depth casual types, \emph{e.g.,} predicting the following audio or visual event based on the previewed scenes, are expected to be explored.

\section{Conclusion}
In this paper, we systematically review and analyze the studies in current audio-visual learning field. The recent advances in this area are categorized into three subjects: audio-visual boosting, cross-modal perception and audio-visual collaboration. To revisit the progress in the current audio-visual learning area from a macro view, a new perspective on audio-visual scene understanding is also provided. We hope this survey can help researchers to get a comprehensive recognition of the audio-visual learning field.


%




\ifCLASSOPTIONcaptionsoff
  \newpage
\fi



\bibliographystyle{IEEEtran}
\bibliography{refer_cleaned}
%

%







\end{document}